\documentclass[lettersize,journal]{IEEEtran}
\usepackage{amsmath,amsfonts}
\usepackage{algorithmic}
\usepackage{algorithm}
\usepackage{array}
\usepackage[caption=false,font=normalsize,labelfont=sf,textfont=sf]{subfig}
\usepackage{textcomp}
\usepackage{stfloats}
\usepackage{url}
\usepackage{verbatim}
\usepackage{graphicx}
\usepackage{cite}
\usepackage{caption}
\usepackage{colortbl}
\usepackage[table]{xcolor}
\usepackage{booktabs}
\usepackage{hyperref}
\hypersetup{
	colorlinks=true,          
	linkcolor=blue,      
	citecolor=blue,            
}

\begin{document}

\title{Spectral Compression Transformer with Line Pose Graph for Monocular 3D Human Pose Estimation}

\author{Zenghao Zheng, Lianping Yang, Hegui Zhu, Mingrui Ye
	
\thanks{Corresponding author: Lianping Yang. 

Zenghao Zheng, Lianping Yang, Hegui Zhu are with the College of Sciences, Northeastern University, Shengyang 110819, China (email: 2300178@stu.neu.edu.cn, yanglp@mail.neu.edu.cn, zhuhegui@mail.neu.edu.cn)
	
Lianping Yang, Hegui Zhu are also with Key Laboratory of Differential Equations and Their Applications, Northeastern University, Liaoning Provincial Department of Education

Mingrui Ye is with Department of Informatics, King's College London (email: mingrui.ye@kcl.ac.uk)}
}

\maketitle

\begin{abstract}
	Transformer-based 3D human pose estimation methods suffer from high computational costs due to the quadratic complexity of self-attention with respect to sequence length. Additionally, pose sequences often contain significant redundancy between frames. However, recent methods typically fail to improve model capacity while effectively eliminating sequence redundancy. In this work, we introduce the Spectral Compression Transformer (SCT) to reduce sequence length and accelerate computation. The SCT encoder treats hidden features between blocks as Temporal Feature Signals (TFS) and applies the Discrete Cosine Transform, a Fourier transform-based technique, to determine the spectral components to be retained. By filtering out certain high-frequency noise components, SCT compresses the sequence length and reduces redundancy. To further enrich the input sequence with prior structural information, we propose the Line Pose Graph (LPG) based on line graph theory. The LPG generates skeletal position information that complements the input 2D joint positions, thereby improving the model's performance. Finally, we design a dual-stream network architecture to effectively model spatial joint relationships and the compressed motion trajectory within the pose sequence. Extensive experiments on two benchmark datasets (i.e., Human3.6M and MPI-INF-3DHP) demonstrate that our model achieves state-of-the-art performance with improved computational efficiency. For example, on the Human3.6M dataset, our method achieves an MPJPE of 37.7mm while maintaining a low computational cost. Furthermore, we perform ablation studies on each module to assess its effectiveness. The code and models will be released.
\end{abstract}

\begin{IEEEkeywords}
Monocular 3D Human Pose Estimation, Transformer, Spectral Compression, Line Graph 
\end{IEEEkeywords}

\section{Introduction}
\IEEEPARstart{3}{D} human pose estimation (HPE) aims to localize 3D body joints from 2D images or video. Monocular 3D human pose estimation is more easily applicable for action recognition \cite{luvizon2020multi}, human-computer interaction \cite{munea2020progress}, and autonomous driving \cite{wiederer2020traffic}. Due to the excellent performance of state-of-the-art (SOTA) 2D pose detectors, most current 3D HPE methods mainly use 2D-to-3D lifting methods. The two-stage approach \cite{cai2019exploiting, chen2021anatomy, wang2020motion} utilizes a lifted model to detect 3D joints from 2D joints detected using an off-the-shelf 2D HPE model.

Recently, transformer-based methods \cite{shan2023diffusion, foo2023unified, li2021tokenpose, chen2023hdformer, zhu2023motionbert, wang2024utilizing} have gained much attention by processing long 2D video pose sequences and spatial joint sequences in 3D HPE. By taking 2D joints coordinates as inputs, these approach regard each video frame of same joint or different joints in the same video frame as a token and leverage long video sequences to capture temporal correlations among consecutive frames. Some methods \cite{xu2020deep,li2020cascaded, peng2024ktpformer} effectively utilize the human body's topological structure to enhance the prior information of pose sequences. By incorporating video pose transformer (VPT) blocks, these methods learn richer pose representations and achieve state-of-the-art results. However, transformer encoder-based approaches have several limitations: (i) The computational complexity of the attention mechanism increases quadratically with sequence length, restricting inference efficiency and the model’s ability to capture long-range dependencies in practical applications. (ii) Video sequences processed by VPT contain substantial redundancy due to the high similarity between adjacent action frames. (iii) The information embedded in pose sequences from video data is inherently limited, and there is no straightforward and effective approach to augment the prior information of the input sequence.

\begin{figure}[ht]
	\centering
	\includegraphics[width=\linewidth]{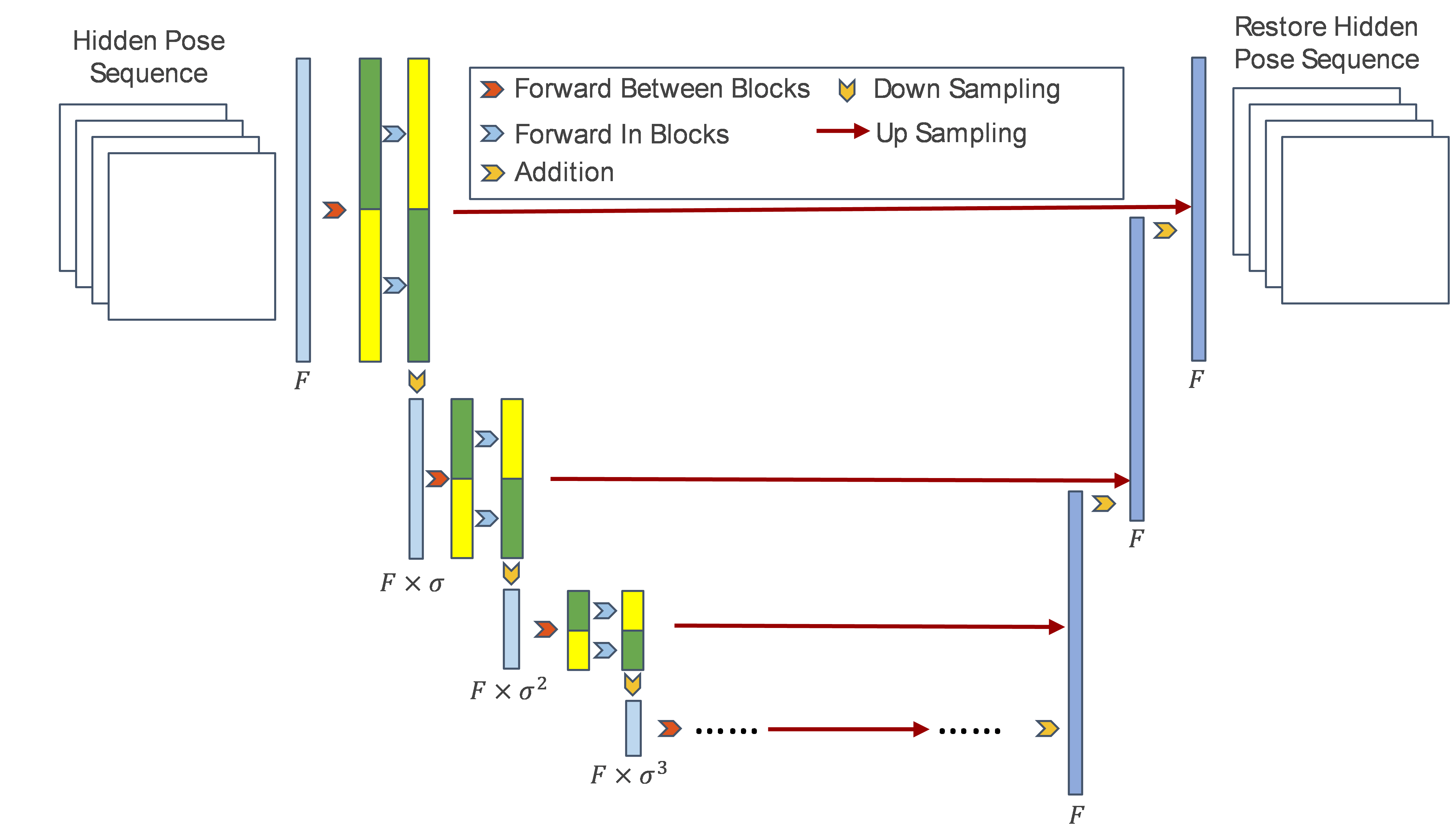}
	\caption{This process involves compressing and upsampling the hidden feature sequence using SCT. Here, $F$ denotes the original sequence frame length. The proposed dual-stream structure progressively downsamples the hidden feature sequence in length while upsampling the output of each layer to reconstruct the hidden pose sequence. The reconstructed sequences from all layers are aggregated to produce the final hidden pose sequence.}
	\label{fig:overview}
\end{figure}

To address the aforementioned challenges, we propose the Spectral Compression Transformer (SCT) to reduce sequence redundancy, thereby improving inference efficiency and reducing memory consumption. While PoseFormerV2\cite{zhao2023poseformerv2} has explored sequence length compression from a frequency-domain perspective, its effectiveness remains constrained. Instead, we introduce a novel approach by compressing sequence redundancy in the hidden features between transformer blocks. Inspired by HoT\cite{li2024hourglass}, we conduct a series of experiments applying the Discrete Cosine Transform (DCT) to the hidden states of VPT blocks and observe that most hidden-layer state information is concentrated in the low-frequency components. By employing a low-pass filter with a frequency-domain compression coefficient $\sigma$, we can effectively compress the temporal pose sequence and reduce redundancy. Building upon prior experimental findings, we propose SCT for sequence compression. Additionally, we introduce a dual-stream architecture to effectively learn compressed sequence features. As illustrated in Figure \ref{fig:overview}, the proposed model employs blocks containing SCT to progressively downsample the latent pose sequence, where the sequence length decreases as the model depth increases. To restore the original temporal resolution, we apply parameter-free upsampling to the hidden features between compressed blocks. The temporal length of each downsampled sequence is recovered via interpolation and subsequently aggregated. As the model depth increases, the pose sequence length is progressively compressed, improving the efficiency of the attention mechanism. Meanwhile, the upsampling process enables precise full-length 3D pose estimation.

Furthermore, drawing inspiration from line graphs in graph theory, we introduce the Line Pose Graph (LPG) to enhance 2D pose priors. In a line graph, each vertex represents an edge in the original graph. Analogously, our approach models each bone as a vertex in the LPG, with edges defined by the joints connecting adjacent bones. By using this vertex-edge interchange method, the LPG is designed to fully represent the bone’s prior information. By constructing the LPG, we generate an enriched set of 2D poses encoding bone joint coordinates. The bone locations and 2D joint positions provide complementary information, reinforcing the priors of the input sequence.

We conduct extensive experiments on two benchmark datasets, Human3.6M\cite{ionescu2013human3} and MPI-INF-3DHP\cite{mehta2017monocular}. The results demonstrate that our proposed method achieves state-of-the-art (SOTA) performance on MPJPE 37.7 mm. To further validate the effectiveness of SCT, we adopt our approach to other mainstream backbone networks, all of which demonstrate a reduction in computational cost while maintaining accuracy. Finally, we conduct ablation studies to evaluate the contribution of each module.

In conclusion, our contributions can be summarized as follows:
\begin{itemize}
	\item{Using a transformer-based encoder model, we propose the Spectral Compression Transformer (SCT) to reduce the sequence length of hidden features between blocks for 3D pose estimation. Furthermore, we design a dual-stream model to effectively learn the compressed features and apply a parameter-free efficient upsampling method to restore the temporal resolution.}
	\item{To obtain richer pose sequence information, we propose the LPG, which interchanges joints and bones to generate new bone coordinates to enhance the 2D prior.}
	\item{The proposed model achieves state-of-the-art results on two benchmark datasets while exhibiting lower computational cost and faster inference speed.}
\end{itemize}

\section{Related Work}
\label{Related work}
\subsection{Transformer-based 3D Human Pose Estimation}
Transformer was first proposed by Vaswani et al. \cite{vaswani2017attention} and showed excellent results because of its strong modeling ability for long-range dependencies. Poseformer \cite{zheng20213d} first introduces the transformer approach to 3D HPE, and subsequently many skeletons begin to be modeled using transformers \cite{qian2023hstformer, wan2021encoder, liu2020attention, lin2021end}. MHFormer \cite{li2022mhformer} generates multiple hypotheses at the pose level and learns their spatio-temporal representations. MixSTE \cite{zhang2022mixste} alternates between spatial and temporal transformer blocks to obtain better temporal-spatial representation. MotionBERT \cite{zhu2023motionbert} proposes a DSTformer to capture the spatio-temporal relationships over the long range of skeletal joints. MotionAGFormer \cite{mehraban2024motionagformer} proposes to use two parallel streams, transformer and GCNFormer, to learn the underlying 3D structure. DualFormer \cite{zhou2023dual} integrates various human contexts and motion details to model long-range dependencies in both joint and limb sequences for effective spatial-temporal modeling.

However, the computational consumption is huge when the the input video’s frame number increases, hence there are some works \cite{tang2023ftcm, zeng2022deciwatch} dedicated to enhance the efficiency of modeling computation. A line \cite{einfalt2023uplift, zhao2023poseformerv2, li2024hourglass} focuses on reducing sequence redundancy in 2D video sequences. Einfalt et al. \cite{einfalt2023uplift} propose a pose uplifting structure to generate temporally sparse 2D pose sequences and produce dense 3D pose sequences. PoseFormerV2 \cite{zhao2023poseformerv2}effectively scales the global receptive fields of 2D joint points using a compressed representation of frequency domain pose sequences. HoT \cite{li2024hourglass} improves the model efficiency by cropping the pose tokens of redundant frames in the intermediate transformer blocks and recovers the full-length pose tokens. Another line of works \cite{tang20233d, li2022exploiting} concentrate on efficient network structures. FTCM \cite{tang2023ftcm} splits pose features into two groups by channel and uses parallel feature-mixing operations to separately model frequency and temporal interactions. STCFormer \cite{tang20233d} accelerates the computation by dividing the input features into two partitions performing temporal and spatial attention simultaneously. Li et al. \cite{li2022exploiting} proposes Strided Transformer Encoder (STE) which uses strided convolutions to shrink the sequence length and aggregate local context information. Previous methods fail to fully explore sequence compression, leading to limited representational capacity. In this work, we investigate the spectral properties of pose sequences from a novel perspective and effectively capture the compressed features.

\subsection{Enhanced Pose Prior}

The input 2D video pose sequences contain a lot of topological information, but feeding the sequences directly into the VPT block does not fully utilize this information. So some methods \cite{xu2020deep, zhang2023learning, wandt2022elepose} preprocess the 2D poses according to the human body topology before feeding them into the backbone network. Xu et al. \cite{xu2020deep} first optimizes the kinematics structure of 2D noise inputs, and then decouple the 2D pose into bone length and bone vector to provide more compact 3D static structure. KTPFormer \cite{peng2024ktpformer} makes full use of the known anatomical structure of the human body and movement trajectories to perform effective global dependency and feature learning on the 2D sequences. Li et al. \cite{li2020cascaded} evolve 2D-3D dataset based on hierarchical human representation and heuristics inspired by prior knowledge. In contrast to prior approaches, we propose the LPG module, a simple yet efficient method that enhances 2D pose priors by leveraging line graph theory.

\subsection{Fourier transform for transformer}
Recently, there have been many applications \cite{qin2021fcanet, csahinucc2022fractional, lee2021fnet} in transformer structures using the Fourier transform. Wang et al. \cite{wang2018packing} relax the computational burden of convolution operations by linearly combining convolutional responses based on discrete cosine transforms. Pan et al. \cite{pan2024dct} use DCT-based attention initialization instead of the traditional initialization strategy. He et al. \cite{he2023fourier} propose Fourier Transformer for hidden sequence to remove redundancies using FFT transform. Li et al. \cite{li2023discrete} propose DCFormer to leverage input information compression in the frequency domain to achieve efficiency-accuracy trade-off. FourierViT \cite{duan2022fourier} introduces a universal attention-to-details architecture for extracting global and local images and invisible features in the frequency domain. DCT-Former \cite{scribano2023dct} fully exploits the properties of DCT to conserve computational memory and reduce inference time. These frequency domain methods are mainly used for cropping and transform characterization. In this work, our proposed SCT module is the first to apply Fourier Attention to 3D human pose estimation and validates the effectiveness of compression in the spectral domain for 3D HPE.

\section{Method}
This section introduces the preliminaries, SCT, LPG module, and network architecture. First, we define the Discrete Cosine Transform (DCT) in the preliminaries and analyze the power spectrum of inter-block hidden features. Our experiments focus on the feature vectors output by each VPT block along the temporal dimension, revealing that most of the energy is concentrated in the low-frequency domain. Inspired by this observation, we propose SCT, which removes sequence redundancy through spectral compression to enhance computational efficiency. Additionally, we introduce the simple yet efficient LPG module to enhance 2D pose representations by incorporating skeletal position information. Finally, we design a dual-stream architecture with SCT to effectively learn the compressed feature vectors and apply a non-parametric recovery method. Figure \ref{fig:backbone} provides an overview of the overall network architecture.

\subsection{Preliminaries}
\label{section:Preliminaries}

\subsubsection{Discrete Cosine Transform}

The Discrete cosine transform (DCT) is a transformation related to Fourier transform. Because DCT exclusively produces real values, it serves as an alternative to Fourier transform within the realm of real numbers. In this paper, we use DCT for the transformation of frequency domain representations since it has been known as the core transform widely used in computer vision tasks \cite{saha2000image, pan2024multichannel}.

Formally, DCT transforms a real number input sequence $x=\{x_0,x_1,\ldots,x_{N-1}\}$ into an output sequence $X=\{X_0,X_1,\ldots,X_{N-1}\}$ of the same length $N$ which we mark it as $\{X_k\}=DCT(\{x_n\})$. The transformation process involves computing each $X_k$ as the inner product of $x$ with a set of cosine basis functions:

\begin{equation}
	X_k=\alpha_k\sum_{n=0}^{N-1}x_n\cos\left(\frac{\pi k(2n+1)}{2N}\right),\quad k=0,1,\ldots,N-1
\end{equation}
where the values of the coefficient $\alpha_k$ are given by:
\begin{equation}
	\alpha_k=\begin{cases}\sqrt{\frac{1}{N}}&\quad if k=0,\\\sqrt{\frac{2}{N}}&\quad if k\neq0\end{cases}
\end{equation}
For a given frequency-domain sequence $X$, compute each $x_n$ using the inverse DCT (IDCT):
\begin{equation}
	x_n=\sum_{k=0}^{N-1}\alpha_kX_k\cos\left(\frac{\pi k(2n+1)}{2N}\right),\quad k=0,1,\ldots,N-1
\end{equation}
which we mark as $\{x_n\}=IDCT(\{X_k\})$.
Discrete Cosine Transform (DCT) can be efficiently computed using the Fast Fourier Transform (FFT), particularly in computer implementations where it significantly speeds up computation.

\subsubsection{Frequency Analysis of Inter-Block Hidden Features}
\label{section:freq_analysis}
The existing 3D HPE networks based on transformers are composed of stacks of autonomously designed blocks. For a specific layer within these blocks, each hidden feature can be regarded as a temporal feature signal (TFS) varying across frames. To analyze the power spectrum of hidden TFS, we perform Fourier transforms along the temporal frame dimension. MixSTE \cite{zhang2022mixste} consists of blocks that alternate in both spatial and temporal domains, yielding promising results. This temporal correlation learning pattern has been adopted by subsequent methods \cite{zhu2023motionbert, mehraban2024motionagformer}, thus we use the pretrained MixSTE backbone network as the experimental subject here. To mitigate interference from individual experimental subjects or specific human actions, we compute the amplitude of intermediate hidden TFS which is obtained by transposing the pre-normalized hidden features output from each temporal block and average across joint and feature dimensions on all Human3.6M \cite{ionescu2013human3} test sets (i.e., S9, S11).

\begin{figure*}[ht]
	\centering
	\includegraphics[width=\linewidth]{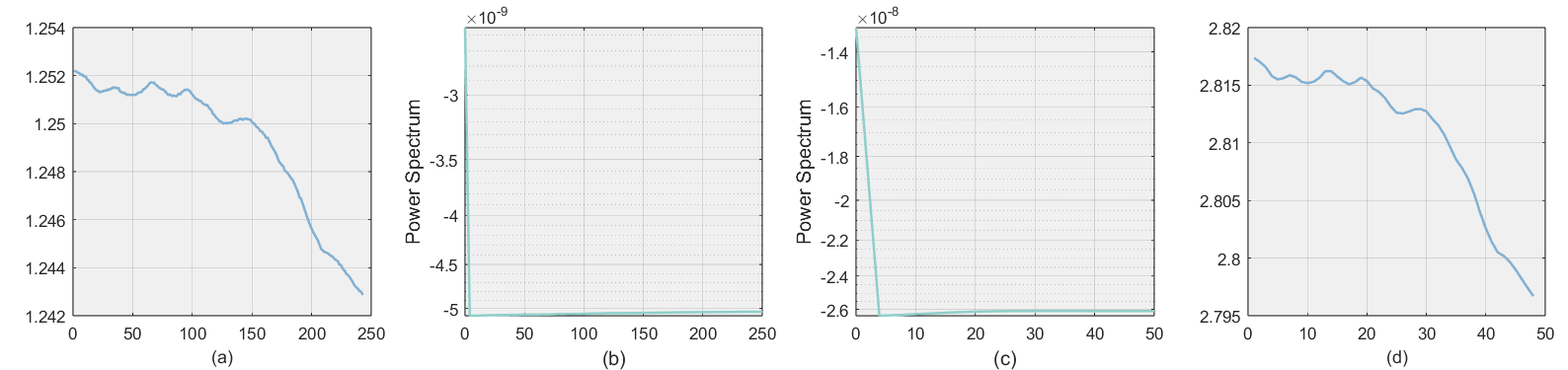}
	\caption{TFS and its associated frequency-domain plots with MixSTE's third block. For subfigures (a) and (d), in the time domain, the horizontal axis represents time, while the vertical axis denotes the signal's magnitude. For the frequency-domain signals in subfigures (b) and (c), the horizontal axis shows the frequency components of the TFS, and the vertical axis displays the power spectral density of the signal at each frequency. Upon examining the images, it becomes evident that the trends of the signals remain largely consistent before and after truncation, with a reduction observed in some jagged noise components.}
	\label{fig:fig_exper}
\end{figure*}

As shown in Figure \ref{fig:fig_exper}, the subfigure (a) depicts the TFS related to time. Upon Fourier transforming these signals, as shown in subplot (b), the energy is predominantly concentrated in the low-frequency components, with the high-frequency components nearly zero. This suggests that when VPT blocks process the relationships between human pose frames, they primarily exploit the low-frequency components of the hidden feature vectors. While high-frequency components preserve fine-grained details of TFS, the high-frequency spectrum exhibits a long tail. Removing unnecessary high-frequency components can effectively reduce sequence redundancy without significantly compromising signal details. Using a low-pass filter to trim high-frequency signals (see subplot (c)), the reconstructed TFS in the time domain effectively preserves original information while filtering out some high-frequency noise (see subplot (d)). Furthermore, the power spectra of hidden features across all blocks is concentrated in the low-frequency domain (see Appendix \ref{Appendix}), suggesting that redundancy across time frames persists as network depth increases. Based on these observations, we argue that performing spectral compression on hidden feature vectors is theoretically justified.

\subsection{Spectral Compression Transformer}
\label{section: SCT}
\begin{figure}
	\centering
	\includegraphics[width=\linewidth]{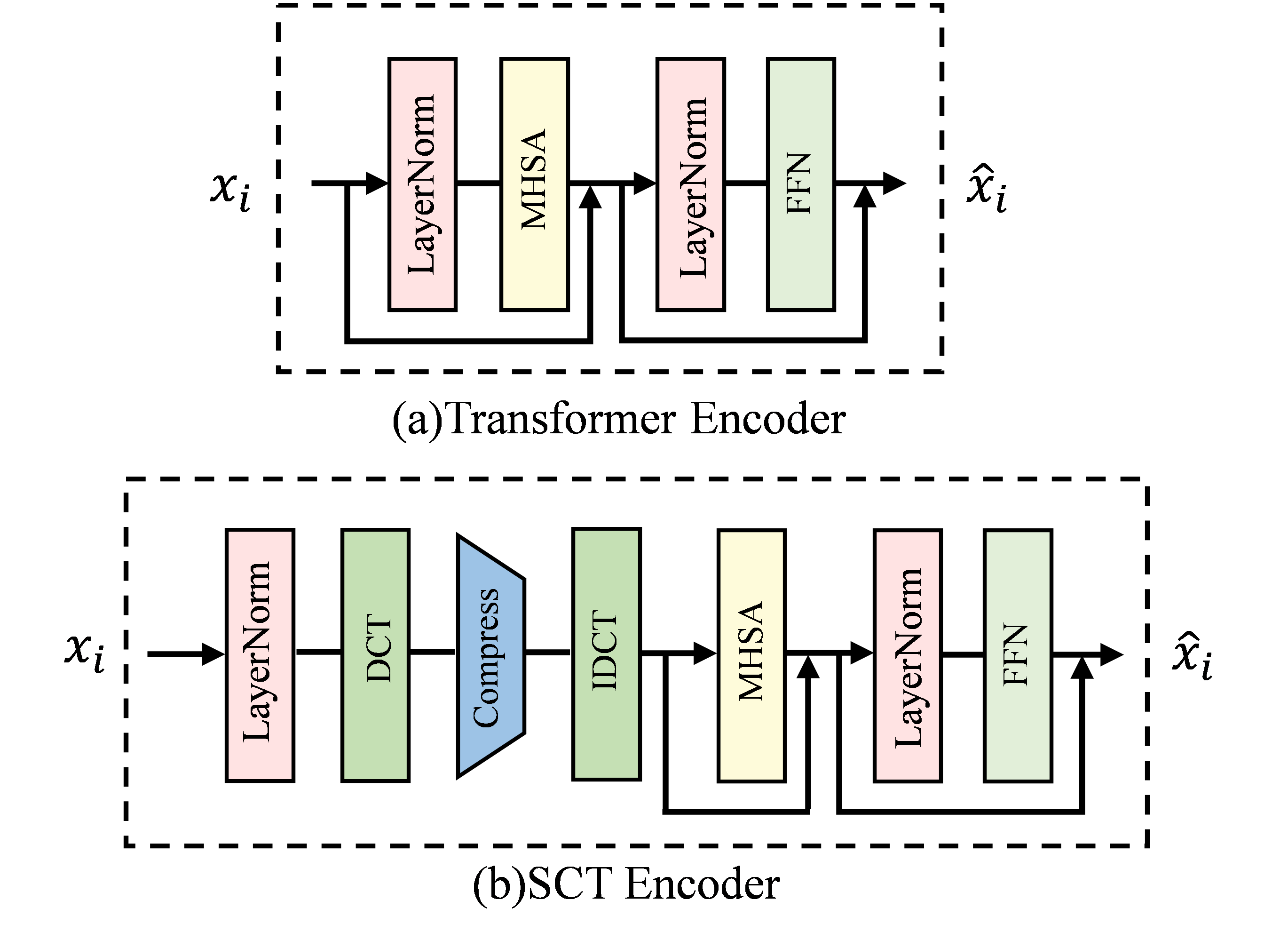}
	\caption{The difference between a standard Transformer encoder and the proposed Spectral Compression Transformer (SCT) encoder, where $Compress$ represents the low-pass filtering process.}
	\label{fig:SCT}
\end{figure}
The SCT module facilitates spectral compression of feature vectors, effectively reducing sequence redundancy and improving computational efficiency. The SCT encoder is a variant of the Transformer encoder, primarily consisting of Fourier transform and its inverse, multi-head self-attention (MHSA), and a feedforward neural network (FFN). As illustrated in Figure \ref{fig:SCT} (a), a standard Transformer encoder block processes an input $x\in\mathbb{R}^{B\times F\times C}$, where $B$ denotes the batch size, $F$ represents the temporal frame length, and $C$ is the embedding dimension. The input $x$ first undergoes layer normalization before being passed into the MHSA, which models dependencies among tokens. In the MHSA mechanism, the input pose sequence is first projected into three different matrices: queries ($Q$), keys ($K$), and values ($V$), which are computed as $Q = xW_Q$, $K = xW_K$, and $V = xW_V$ where $W_Q$, $W_K$, and $W_V$ are learned projection matrices. The attention scores between each query and all keys are then computed using the dot-product attention formula:
\begin{equation}
	\text{Attention}(Q,K,V)=\text{softmax}\left(\frac{QK^T}{\sqrt{d_k}}\right)V
\end{equation}
where $d_k$ is the dimensionality of the keys. Then, $Q$, $K$, and $V$ are split into $H$ heads along the feature dimension. Specifically, for each head $i$, the attention output is calculated independently as $\text{head}_i = \text{Attention}(Q_i, K_i, V_i)$. The outputs of all heads are then concatenated and projected back into the original feature space:
\begin{equation}
	\text{MHSA}(Q,K,V)=\text{Concat}(\text{head}_1,\text{head}_2,\ldots,\text{head}_h)W_O
\end{equation}
where $W_O$ is a learned projection matrix, and $h$ is the number of attention heads.
Next, a feedforward neural network (FFN) is employed to further process the features:
\begin{equation}
	\text{FFN}(x_m)=gelu(x_mW_1+b_1)W_2+b_2
\end{equation}
where $gelu$ represents the GELU activation function, $x_m$ serves as the input to the FFN layer, and $W_1$ and $W_2$, as well as $b_1$ and $b_2$ denote the learnable weight matrices and biases, respectively.

As illustrated in Figure \ref{fig:SCT} (b), the SCT module differs from a standard Transformer encoder primarily in the preprocessing of the input before the attention mechanism. To be specific, given an input hidden sequence $x\in\mathbb{R}^{F\times J\times C}$ with $F$ frames, $J$ joints and $C$ feature dimensions, we first perform layer normalization on the hidden sequence, and then permute its dimensions to interpret it as signals $s \in \mathbb{R}^{J \times C \times F}$ for each feature across time frames for each joint. We perform a DCT on $\{s_n\},0<n<F,s_n\in\mathbb{R}^{J\times C}$ along the time dimension to obtain $\{S_k\}$:
\begin{equation}
	\{S_k\}=DCT(\{s_n\}),0<k<F-1
\end{equation}
where $S_k\in\mathbb{R}^{J\times C}$. Then, by defining a truncation coefficient $\sigma \in(0,1)$, we apply a low-pass filter to $\{S_k\}$, obtaining $\{S_{k^{\prime}}\}$:
\begin{equation}
	\{S_{k^{\prime}}\} = \textit{low-pass}(\{S_k\}),0<k<F-1,0<k^{\prime}<f-1
\end{equation}
where $f=\left\lceil{F\times{\sigma}}\right\rceil$ and $\sigma$ is an important parameter that balances the model's learning of temporal frame correlations and computational speed. Thus, we obtain the low-frequency information representing the global temporal frame characteristics, while filtering out high-frequency noise. Finally, the resulting shorter frequency-domain sequence $\{S_{k^{\prime}}\}$ is transformed back into the time domain features along the time dimension $\{\hat{s}_n\}$ using IDCT:
\begin{equation}
	\{\hat{s}_n\} = IDCT(\{S_{k^{\prime}}\}),0<n<f-1
\end{equation}
where $\hat{s}_n\in\mathbb{R}^{J\times C}$. The features and time dimensions of TFS $\{\hat{s}_n\}$ are changed to obtain the compressed features $\hat{x}\in\mathbb{R}^{J\times f\times C}$. The transformed input $\hat{x}$ is passed to the subsequent MHSA and FFN layers. Similarly, we apply a residual connection to MHSA; however, it is important to note that the residual connection operates directly on the compressed sequence. The computational process of the module can be expressed as:
\begin{equation}
	\label{func:SCT}
	\begin{aligned}\hat{x} &= IDCT(low\mbox{-}pass(DCT(\text{LN}(x)))) \\
		\hat{x}^{\prime} &= \hat{x} + \text{MSA}(\hat{x})	\\
		\hat{x}^{t} &= \hat{x}^{\prime} + \text{FFN}(\text{LN}(\hat{x}^{\prime}))\end{aligned}
\end{equation}
where \text{LN}(·) is the LayerNorm layer, \text{MSA}(·) stands for Multi-Head Self-Attention, and \text{FFN}(·) refers to a feedforward neural network. In the above equation, when performing the DCT, the transposition along the $x$ dimension is omitted.

Here, we analyze the time complexity of the module. A standard transformer layer typically consists of a multi-head self-attention (MSA) mechanism and a feed-forward layer. For an input sequence of length $N$ and hidden dimensionality $D$, the time complexity is $\mathcal{O}\left(ND^2+N^2D\right)$. SCT allows for compressing the sequence length from $N$ to ${\sigma}N$ within $\mathcal{O}\left(N\log N\right)$ time complexity which is negligible. Thus, the time complexity can be reduced to $\mathcal{O}\left({\sigma}ND^2+{\sigma}^2N^2D\right)$. As the SCT-based network deepens, the sequence length is progressively compressed. The time complexity of an N-layer network with SCT encoders is given by
\begin{equation}
	\mathcal{O}(\sum_{i=1}^N(\sigma^{i}ND^2+{\sigma}^{2i}N^2D)).
\end{equation}
However, the time complexity of a multi-layer standard Transformer encoder scales linearly with the number of layers, reaching N times the complexity of a single layer.

\subsection{Line Pose Graph}
\begin{figure}[ht]
	\centering
	\includegraphics[width=\linewidth]{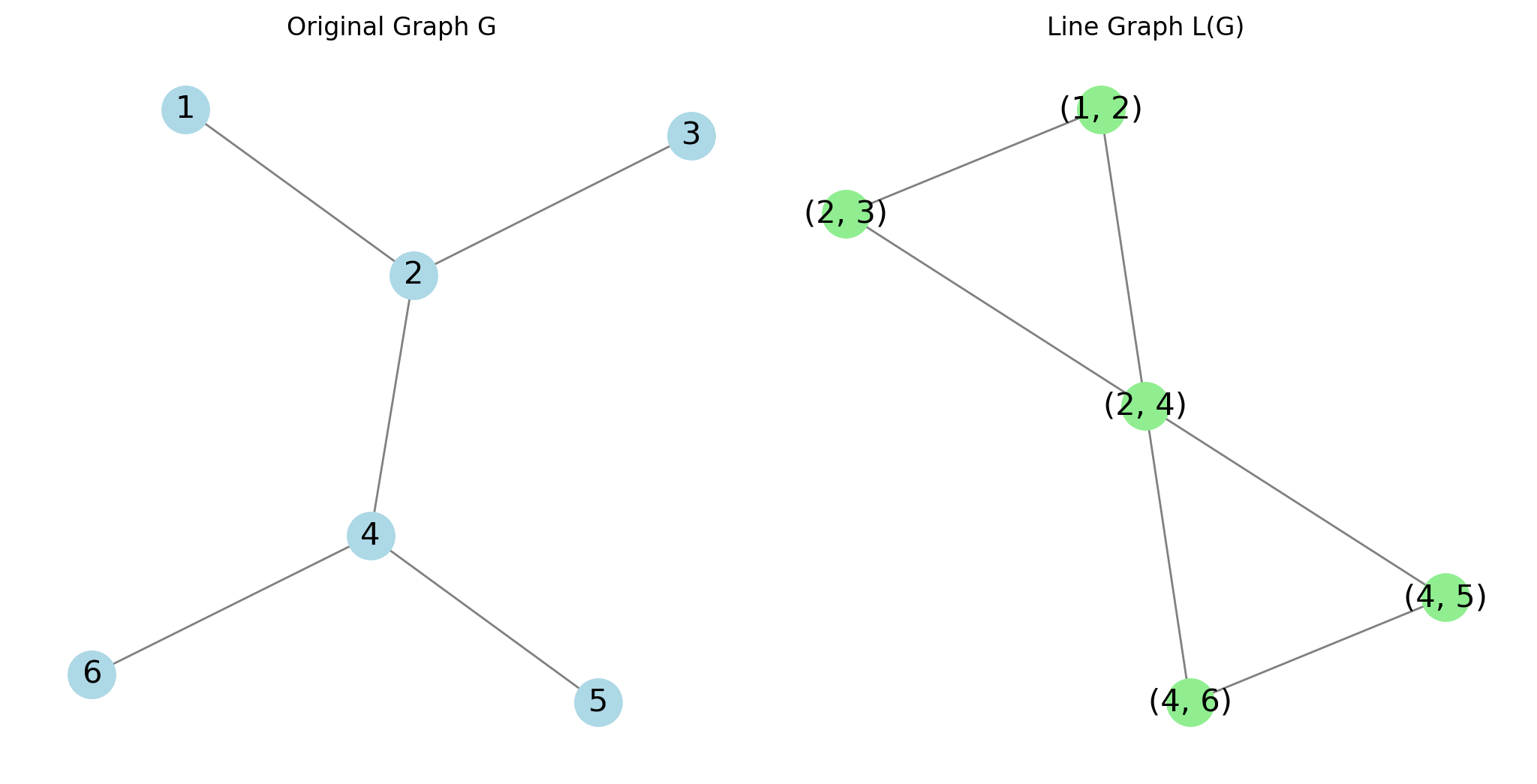}
	\caption{A case of transformation from an original graph $G$ to its line graph $L(G)$.}
	\label{fig:linegraph}
\end{figure}
\begin{figure}[ht]
	\centering
	\includegraphics[width=\linewidth]{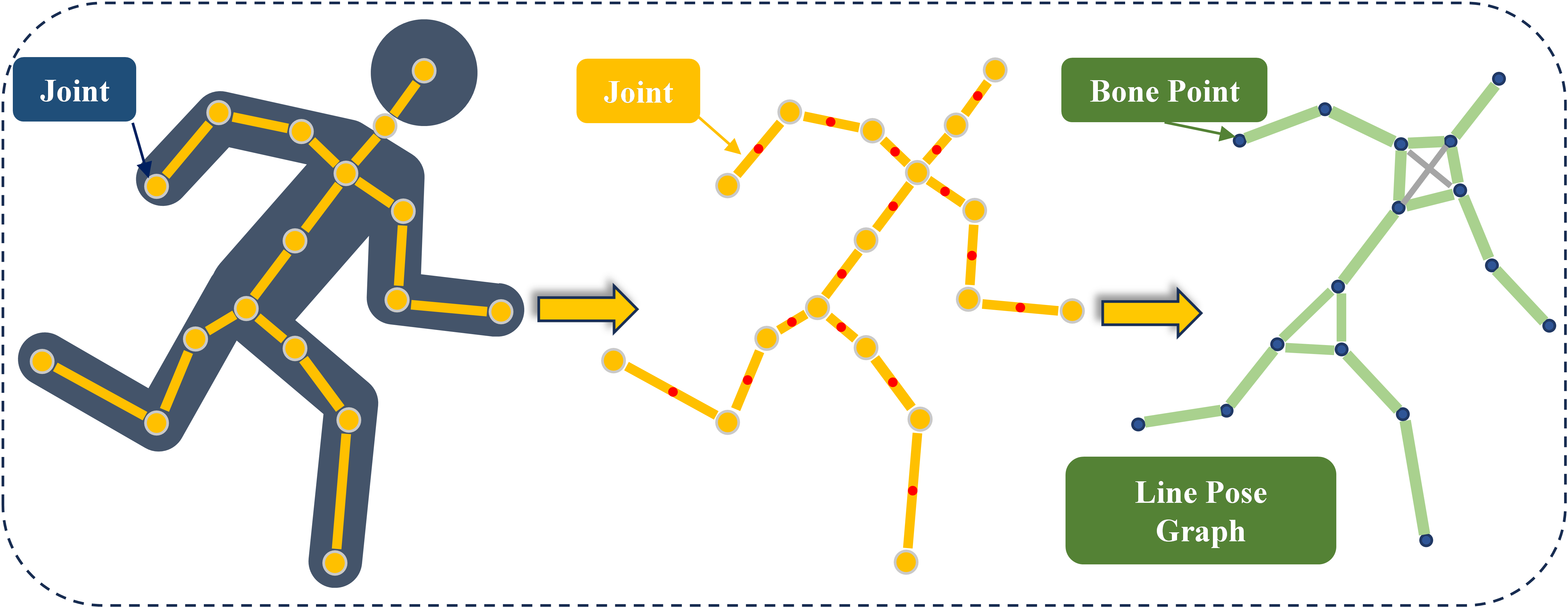}
	\caption{Illustration of the transformation from the original pose to the line pose. Yellow dots represent joints, and yellow lines connecting these joints represent bones. Post-transformation, the red dots represent the midpoints of the bones, and in the line pose, there are edges between the vertices corresponding to bones that share a common joint in the original pose.}
	\label{fig:linepose}
\end{figure}
A common approach in previous models is to use 2D pose sequences as input. However, 2D joints provide only limited prior information to the model. While earlier methods \cite{xu2020deep, cai2023htnet, cai2024disentangled} have adjusted the model input by analyzing the human body topology, they have not examined the problem from the perspective of 2D poses and typically rely on complex network architectures to process this information. We propose the Line Pose Graph (LPG) module to enrich the input 2D pose representation. LPG is a straightforward and efficient module that generates a new set of human body pose topology keypoints. 

The inspiration for LPG comes from the concept of line-graph \cite{chartrand1969connectivity} in graph theory. The line graph corresponding to a graph $G$ is a graph that reflects the adjacency of the edges in $G$, denoted as $L(G)$. As shown in Figure \ref{fig:linegraph}, simply put, $L(G)$ abstracts each edge in $G$ into a vertex; if two edges in the original graph are adjacent, then a corresponding edge is drawn between their respective vertices in the line graph, i.e., $G=(V,E),L(G)=(E,\tilde{E}),\text{and}(e_1,e_2)\in\tilde{E}\iff e_1\cap e_2\neq\emptyset$. The number of vertices in the line graph equals the number of edges in the original graph. By considering the joints and bones in a 2D pose as vertices and edges, respectively, a 2D pose can be viewed as an acyclic graph. We transform the 2D pose graph into an LPG by following the transformation of a line graph. Any point on each edge in the pose graph $P$ is selected as a vertex of the line pose $L(P)$, and the vertices connecting two edges are treated as edges of $L(P)$. The line pose contains different information from the original pose, including the information of bone positions, which can supplement the topological relationship between adjacent joints. When estimating 3D poses, the model not only considers the joint information but also uses bone positions information to constrain the predicted joints, preventing them from aligning in incorrect directions. Figure \ref{fig:linepose} illustrates our Line Pose. Specifically, given 2D pose coordinates $\{J_i\},J_i\in\mathbb{R}^{2},i=0,\ldots,16$ with 17 joints, We use the mean of the parent joint $J_p$ and child joint $J_c$ coordinates to represent the position of the bone $B_i$:
\begin{equation}
	\label{function:LPG}
	B_i = \frac{J_{i^{p}}+J_{i^{c}}}{2}
\end{equation}
where $J_{i^{p}}$ and $J_{i^{c}}$ denote the parent joint and child joint of the $i$th bone, respectively.
To facilitate integration with the original pose, we compute the positions $\{B_i\},B_i\in\mathbb{R}^{2},i=0,\ldots,15$ of 16 bones connecting all joints, combine them with the root joint $J_0$, and thereby form a new set $\{B_i\},i=0,\ldots,16$ of skeletal coordinates for the human body. LPG computes the skeletal joint positions for each time frame in the input 2D pose sequence. When information from certain joints is lost or affected by noise, the bone position information can help recover or correct the estimation of these joints, thereby enhancing the model's robustness.

\subsection{Network Architecture}
\label{section: network}
\begin{figure*}[ht]
	\centering
	\includegraphics[width=\linewidth]{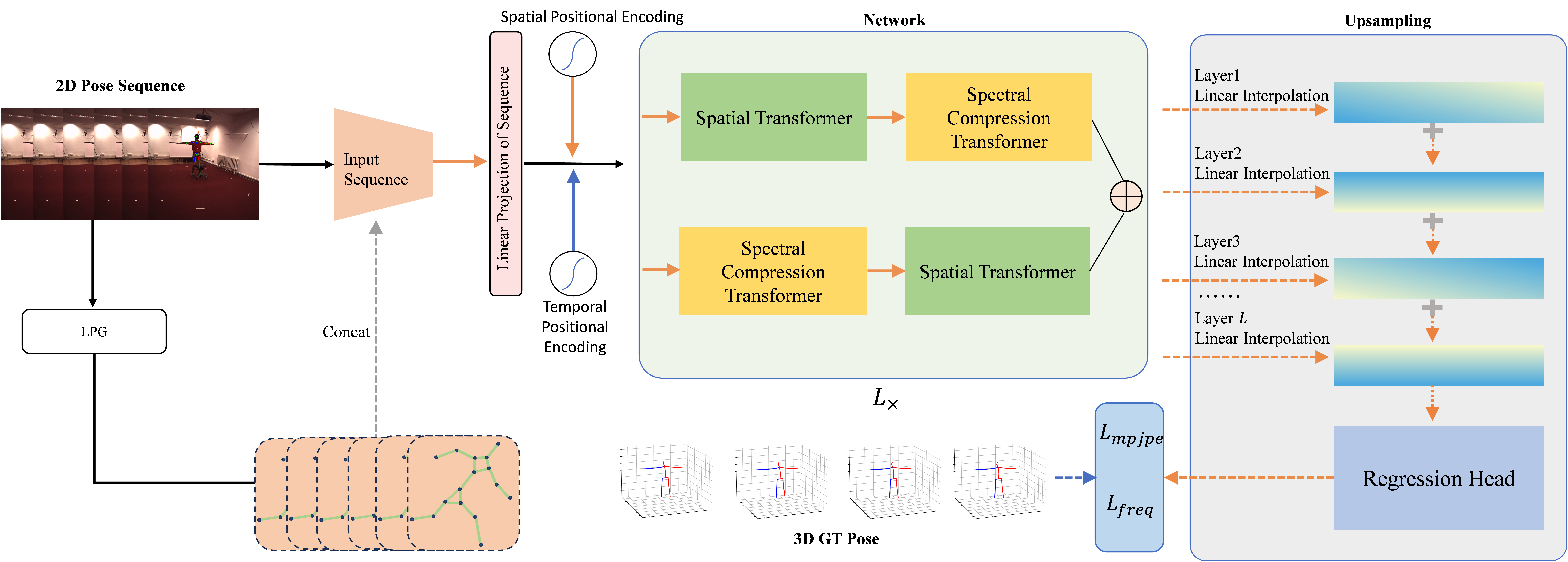}
	\caption{The overall network architecture of the proposed model. The 2D pose sequence coordinates are processed through the Line Pose Graph (LPG) module to generate bone position coordinates. These coordinates are concatenated with the 2D pose sequence along the channel dimension to form the input sequence. The input sequence first undergoes linear projection for feature embedding, followed by the addition of spatial and temporal positional encodings. This embedded sequence is then processed through our network architecture, where each hierarchical layer employs SCT to progressively reduce the sequence length. The adaptive fused features from dual processing branches are propagated to subsequent layers, with parallel feature copies undergoing linear interpolation-based upsampling. Multi-layer upsampled features are aggregated and fed into the regression head to predict the 3D pose sequence.}
	\label{fig:backbone}
\end{figure*}
We design a dual-stream network architecture that incorporates the proposed SCT and LPG modules while restoring temporal resolution. Building on insights from \cite{zhang2022mixste, zhu2023motionbert}, we recognize the strong performance of spatiotemporal Transformer architectures in 3D human pose estimation tasks, as they demonstrate robust capabilities in modeling both temporal and spatial information in pose sequences. The parallel architecture design facilitates a more comprehensive understanding of spatiotemporal dynamics. The overall model architecture is illustrated in Figure \ref{fig:backbone}.

For a given input 2D pose sequence $x$, the corresponding bone coordinates $b$ are computed from Equation \ref{function:LPG}. Then, $x$ and $b$ are concatenated along the final dimension and passed through the embedding layer, where they are mapped to a high-dimensional feature space:
\begin{equation}
	X = \text{Concat}(x, b)W_E
\end{equation}
where \textit{Concat}(·) represents stitching along the joint coordinate dimension, $W_E$ a linear projection matrix. $X$ needs to be augmented with temporal positional encoding $E_{temp}$ and spatial positional encoding $E_{pos}$ to retain spatiotemporal information. Next, we feed the embedded features into our proposed dual-stream network block. The basic block of the model consists of two branches, each employing SCT for sequence compression. For the first branch, the model initially learns the dependencies between different joints:
\begin{equation}
	\begin{aligned}
		X^{\prime}_i &= X_i + S\mbox{-}MHSA(LN(X_i)) \\
		X^{\prime\prime}_i &= X^{\prime}_i+FFN(LN(X^{\prime}_i))
	\end{aligned}
\end{equation}
where $i$ represents the $i$-th block, with $i$ starting from 0, $X_i$ denotes the input to the current block, $S\mbox{-}MHSA$ performs self-attention over spatial tokens and $X^{\prime\prime}_i\in\mathbb{R}^{b\times f_i\times j\times c}$ represents the output of the spatial encoder. Then, the temporal and spatial dimensions are swapped, yielding $\hat{X}\in\mathbb{R}^{b\times j\times f_i\times c}$, which is then fed into the SCT encoder:
\begin{equation}
	\hat{X}^{\prime\prime}_i = SCT(\hat{X})
\end{equation}
where $SCT$ represents the process of Equation \ref{func:SCT} and $\hat{X}^{\prime\prime}\in\mathbb{R}^{b\times (\sigma f_i)\times j\times c}$ represents the output of the SCT encoder. The SCT module processes the feature sequence that has been thoroughly learned by the spatial encoder.
For the other branch, the SCT module is applied first, followed by the spatial encoder block:
\begin{equation}
	\begin{aligned}
		\check{X}_i &= SCT(X_i)\\
		\check{X}^{\prime}_i &= \check{X}_i + S\mbox{-}MHSA(LN(\check{X}_i)) \\
		\check{X}^{\prime\prime}_i &= \check{X}^{\prime}_i+FFN(LN(\check{X}^{\prime}_i))
	\end{aligned}
\end{equation}
where $\check{X}^{\prime\prime}_i \in \mathbb{R}^{b \times (\sigma f_i) \times j \times c}$ denotes the output of this branch. SCT first eliminates sequence redundancy, then spatial attention is used to capture the pose features of the compressed sequence, forming a complement to the first branch. Finally, the adaptive fusion method \cite{zhu2023motionbert} is employed to compute the combined output $X_{i+1}$ of the two branches.

To restore the full sequence length, we design a parameter-free upsampling method, as shown in Figure \ref{fig:backbone}. There are various upsampling techniques, and we choose interpolation because it requires no additional parameters and produces optimal results. The time resolution restoration process is applied to the hidden features between each block. For the hidden feature $X_i\in\mathbb{R}^{b\times f_i\times j\times c},i\in{l},0<f_i\leq{F}$ outputted by the $i$-th block, where $l$ denotes the total number of layers and $f_i$ depends on the current layer $i$, we employ linear interpolation for upsampling along the temporal dimension. Subsequently, we obtain the complete hidden sequence $H_i\in\mathbb{R}^{b\times F\times j\times c}$ for each layer, and we sum the hidden features across all layers and feed the result into a regression head to predict the 3D pose sequence:
\begin{equation}
	O = (\sum_{i=0}^{l}H_i)W_o
\end{equation}
where $O\in\mathbb{R}^{b\times F\times j\times 3}$, $W_o\in \mathbb{R}^{c\times 3}$ is a projection matrix used to generate the final 3D pose. 
Summing the outputs of all layers integrates the rich semantic information from shallow layers and the refined trajectory features from deep layers, facilitating the accurate prediction of the final 3D pose.

For the loss function, we follow previous works \cite{zhang2022mixste, li2024hourglass} and adopt the commonly used loss function MPJPE. Additionally, we introduce a frequency-domain loss function $\mathcal{L}_{freq}$, which substantially reduces the difficulty of extracting frequency-domain information. The final loss function can be represented as:
\begin{equation}
	\begin{aligned}
		\mathcal{L}_{freq}&=\frac{1}{TJ} \sum_{t=1}^{f} \sum_{j=1}^{J}\|DCT(pred_{t,j})-DCT(gt_{t,j})\|_2      \\
		\mathcal{L}_{3D} &= \mathcal{L}_{mpjpe}+\lambda\mathcal{L}_{freq}
	\end{aligned}
\end{equation}
where $\lambda$ is a weighting coefficient of frequency domain loss, $pred_{t,j}, gt_{t,j}$ respectively denotes the predicted position and the ground truth.

\section{Experiments}

\subsection{Dataset and evaluation metric}

We evaluate our method on two 3D HPE bench-mark datasets: Human3.6M \cite{ionescu2013human3} and MPI-INF-3DHP \cite{mehta2017monocular}. 

\text{Human3.6M} is an indoor scenes dataset with 3.6 million video frames. It has 11 professional actors, performing 15 actions under 4 synchronized camera views. We follow the previous work \cite{zhao2023poseformerv2, pavllo:videopose3d:2019}, using 5 subjects for training (S1, S5, S6, S7, S8) and 2 subjects for testing (S9, S11). For evaluation metrics, following previous work \cite{zhang2022mixste, pavllo:videopose3d:2019}, we employ two protocols to assess model performance. Protocol 1 uses MPJPE, which measures the average Euclidean distance between predicted and ground truth 3D keypoint coordinates in millimeters. Protocol 2 employs procrustes MPJPE (P-MPJPE), which refers to the reconstruction error after the predicted 3D pose is aligned to the ground-truth 3D pose using procrustes analysis.

\text{MPI-INF-3DHP} is another popular 3D HPE dataset. This dataset comprises 1.3 million frames collected from either indoor constrained scenes or outdoor complex environments. Following the setup of \cite{zhang2022mixste, mehraban2024motionagformer}, we employed MPJPE, percentage if correct keypoint (PCK) with the threshold of 150mm, and area under curve (AUC) as evaluation metrics.

\subsection{Implementation Details}
The experiments are conducted on an Ubuntu 20.04 system, utilizing the PyTorch framework and an RTX 4090 GPU with 24GB of memory. We use 2D poses with confidence scores detected by the Stacked Hourglass (SH) network \cite{newell2016stacked} as input. Additionally, ground-truth 2D poses are also used as input to assess the upper performance limit of the model. The model consists of 5 layers, with a feature dimension $c$ of 512. The input sequence has 243 frames, each containing 17 joints. During training, the batch size is set to 16, and training is conducted for 120 epochs. The models are optimized using AdamW \cite{loshchilov2017decoupled} with a weight decay of 0.01. The learning rate, learning rate decay, and dropout rate are set to \(2 \times 10^{-4}\), \(0.99\), and \(0.1\), respectively. Pose-level augmentation via horizontal flipping is applied during training. For two key hyperparameters, the compression coefficient $\sigma$ is set to 0.6, and the weight of the frequency-domain loss function $\lambda$ is set to 1. 

\subsection{Comparison with State-of-the-art Methods}

\begin{table*}[ht]\LARGE
	\caption{Quantitative comparisons of 3D human pose estimation per action on Human3.6M. MPJPE~(mm) and P-MPJPE~(mm) using detected 2D pose sequence. $T$ denotes the number of input frames. (*) denotes using HRNet~\cite{sun2019deep} for 2D pose estimation. The best results are highlighted in bold, and the second-best results are underlined.}
	\centering
	\resizebox{\linewidth}{!}{
		\begin{tabular}{lc|ccccccccccccccc|c}
			\hline
			\textbf{MPJPE} 											& $T$ & Dire. & Disc. & Eat & Greet & Phone & Photo & Pose & Purch. & Sit & SitD & Smoke & Wait & WalkD & Walk & WalkT & Avg\\
			\hline
			*MHFormer~\cite{li2022mhformer}CVPR'22					& 351 & 39.2 & 43.1 & 40.1 & 40.9 & 44.9 & 51.2 & 40.6 & 41.3 & 53.5 & 60.3 & 43.7 & 41.1 & 43.8 & 29.8 & 30.6 & 43.0\\
			MixSTE~\cite{zhang2022mixste}CVPR'22 					& 243 & 37.6 & 40.9 & 37.3 & 39.7 & 42.3 & 49.9 & 40.1 & 39.8 & 51.7 & 55.0 & 42.1 & 39.8 & 41.0 & 27.9 & 27.9 & 40.9\\
			P-STMO~\cite{shan2022p}ECCV'22							& 243 & 38.9 & 42.7 & 40.4 & 41.1 & 45.6 & 49.7 & 40.9 & 39.9 & 55.5 & 59.4 & 44.9 & 42.2 & 42.7 & 29.4 & 29.4 & 42.8\\
			StridedFormer~\cite{li2022exploiting}TMM'23				& 351 & 40.3 & 43.3 & 40.2 & 42.3 & 45.6 & 52.3 & 41.8 & 40.5 & 55.9 & 60.6 & 44.2 & 43.0 & 44.2 & 30.0 & 30.2 & 43.7\\
			Einfalt~\textit{et al.}~\cite{einfalt2023uplift}WACV'23	& 351 & 39.6 & 43.8 & 40.2 & 42.4 & 46.5 & 53.9 & 42.3 & 42.5 & 55.7 & 62.3 & 45.1 & 43.0 & 44.7 & 30.1 & 30.8 & 44.2\\
			STCFormer~\cite{tang20233d}CVPR'23						& 243 & 39.6 & 41.6 & 37.4 & 38.8 & 43.1 & 51.1 & 39.1 & 39.7 & 51.4 & 57.4 & 41.8 & 38.5 & 40.7 & 27.1 & 28.6 & 41.0\\
			STCFormer-L~\cite{tang20233d}CVPR'23					& 243 & 38.4 & 41.2 & 36.8 & 38.0 & 42.7 & 50.5 & 38.7 & 38.2 & 52.5 & 56.8 & 41.8 & 38.4 & 40.2 & \underline{26.2} & 27.7 & 40.5\\
			UPS~\cite{foo2023unified}CVPR'23						& 243 & 37.5 & 39.2 & 36.9 & 40.6 & \textbf{39.3} & \textbf{46.8} & 39.0 & 41.7 & 50.6 & 63.5 & 40.4 & 37.8 & 44.2 & 26.7 & 29.1 & 40.8\\
			GLA-GCN~\cite{yu2023gla}ICCV'23                 &243  &41.3 &44.3 &40.8 &41.8 &45.9 &54.1 &42.1 &41.5 &57.8 &62.9 &45.0 &42.8 &45.9 &29.4 &29.9 &44.4\\
			MotionBERT~\cite{zhu2023motionbert}CVPR'23				& 243 & \underline{36.6} & 39.3 & 37.8 & 33.5 & 41.4 & 49.9 & \underline{37.0} & 35.5 & 50.4 & 56.5 & 41.4 & 38.2 &37.3 & \underline{26.2} & \underline{26.9} & 39.2\\
			HDFormer~\cite{chen2023hdformer} IJCAI'23				& 96 & 38.1 & 43.1 & 39.3 & 39.4 & 44.3 & 49.1 & 41.3 & 40.8 & 53.1 & 62.1 & 43.3 & 41.8 & 43.1 & 31.0 & 29.7 & 42.6\\
			MotionAGFormer~\cite{mehraban2024motionagformer}WACV'24 & 243 &36.8 &\underline{38.5} &\underline{35.9} &\underline{33.0} &41.1 &48.6 &38.0 &\underline{34.8} &\textbf{49.0} &\textbf{51.4} &\underline{40.3} &\underline{37.4} &\underline{36.3} &27.2 &27.2 &\underline{38.4} \\
			HoT~\cite{li2024hourglass} CVPR'24					    & 243 &-&-&-&-&-&-&-&-&-&-&-&-&-&-&-&39.0 \\
			FTCM~\cite{tang2023ftcm}TCSVT'24						& 351 & 42.2 & 44.4 & 42.4 & 42.4 & 47.7 & 55.8	& 42.7 & 41.9 & 58.7 & 64.5 & 46.1 & 44.2 & 45.2 & 30.6 & 31.1 & 45.3\\
			FPMT~\cite{zhong2023frame}TMM'24						& 243 &36.9 &39.6 &36.9 &39.3 &41.8 &48.3 &38.4 &38.7 &51.1 &53.7 &41.9 &38.7 &40.4 &27.7 &27.9 &40.1 \\
			\hline
			\rowcolor{gray!20} Ours     							& 243  & \textbf{35.8} & \textbf{38.3} & \textbf{35.7} & \textbf{32.2} & \underline{39.7} & \underline{47.3} & \textbf{36.0} & \textbf{34.5} & \underline{50.0} & \underline{52.9} & \textbf{39.4} &\textbf{36.8}  & \textbf{35.7} & \textbf{25.2} & \textbf{25.3} & \textbf{37.7}\\
			
			\hline
			\hline
			\textbf{P-MPJPE} 					  				  & $T$ & Dire.& Disc. & Eat & Greet & Phone & Photo & Pose & Purch. & Sit & SitD & Smoke & Wait & WalkD & Walk & WalkT & Avg\\
			*MHFormer~\cite{li2022mhformer}CVPR'22 				  & 351 & 31.5 & 34.9 & 32.8 & 33.6 & 35.3 & 39.6 & 32.0 & 32.2 & 43.5 & 48.7 & 36.4 & 32.6 & 34.3 & 23.9 & 25.1 & 34.4\\
			MixSTE~\cite{zhang2022mixste}CVPR'22  				  & 243 & 30.8 & 33.1 & \textbf{30.3} & 31.8 & \textbf{33.1} & 39.1 & 31.1 & 30.5 & 42.5 & \textbf{44.5} & \textbf{34.0} & \underline{30.8} & 32.7 & 22.1 & \underline{22.9} & 32.6\\
			P-STMO~\cite{shan2022p}ECCV'22 		  				  & 243 & 31.3 & 35.2 & 32.9 & 33.9 & 35.4 & 39.3 & 32.5 & 31.5 & 44.6 & 48.2 & 36.3 & 32.9 & 34.4 & 23.8 & 23.9 & 34.4\\
			GLA-GCN~\cite{yu2023gla}ICCV'23						  &243  &32.4 &35.3 &32.6 &34.2 &35.0 &42.1 &32.1 &31.9 &45.5 &49.5 &36.1 &32.4 &35.6 &23.5 &24.7 &34.8 \\
			StridedFormer~\cite{li2022exploiting}TMM'23			  & 351 & 32.7 & 35.5 & 32.5 & 35.4 & 35.9 & 41.6 & 33.0 & 31.9 & 45.1 & 50.1 & 36.3 & 33.5 & 35.1 & 23.9 & 25.0 & 35.2\\
			Einfalt~\textit{et al.}~\cite{einfalt2023uplift}WACV'23& 351 & 32.7 & 36.1 & 33.4 & 36.0 & 36.1 & 42.0 & 33.3 & 33.1 & 45.4 & 50.7 & 37.0 & 34.1 & 35.9 & 24.4 & 25.4 & 35.7\\
			MotionBERT~\cite{zhu2023motionbert}CVPR'23	 		  & 243 & 30.8 &32.8 & 32.4 &28.7 & 34.3 & 38.9  & \underline{30.1} & \underline{30.0} & 42.5 & 49.7 & 36.0 & \underline{30.8} & 31.7 & \underline{22.0} & 23.0 & 32.9\\
			HDFormer~\cite{chen2023hdformer}IJCAI'23     		  & 96   & \textbf{29.6} & 33.8 & 31.7 & 31.3 & \underline{33.7} & \underline{37.7} & 30.6 & 31.0 & \textbf{41.4} & 47.6 & 35.0 & 30.9 & 33.7 & 25.3 & 23.6 & 33.1\\
			MotionAGFormer~\cite{mehraban2024motionagformer}WACV'24 &243 &31.0 &\underline{32.6} &31.0 &\underline{27.9} &34.0 &38.7 &31.5 &\underline{30.0} &\textbf{41.4} &\underline{45.4} &34.8 &\underline{30.8} &\underline{31.3} &22.8 &23.2 &\underline{32.5} \\
			FTCM~\cite{tang2023ftcm}TCSVT'24	   				  & 351 & 31.9 & 35.1 & 34.0 & 34.2 & 36.0 & 42.1 & 32.3 &31.2 & 46.6 & 51.9 & 36.5 & 33.8 & 34.4 & 24.0 & 24.9 & 35.3\\
			\hline
			\rowcolor{gray!20}Ours     	  						  & 243  & \underline{29.7} & \textbf{31.6} & \underline{30.6} & \textbf{27.2} & \textbf{33.1} & \textbf{37.6} & \textbf{29.6} & \textbf{29.9} & \underline{41.6} & 46.6 & \underline{34.5} & \textbf{29.9} & \textbf{30.6} & \textbf{21.4} & \textbf{21.8} & \textbf{31.7} \\
			\hline
		\end{tabular}
	}
	\label{tab:human3.6m-comparison-action}
\end{table*}

\begin{table*}[ht]\LARGE
	\caption{Quantitative comparisons of 3D human pose estimation per action using 2D GroundTruth (GT) human poses on Human3.6M. $T$ denotes the number of input frames. The best results are highlighted in bold, and the second-best results are underlined.}
	\centering
	\resizebox{\linewidth}{!}{
		\begin{tabular}{lc|ccccccccccccccc|c}
			\hline
			\textbf{MPJPE} 											& $T$ & Dire. & Disc. & Eat & Greet & Phone & Photo & Pose & Purch. & Sit & SitD & Smoke & Wait & WalkD & Walk & WalkT & Avg\\
			\hline
			Poseformer~\cite{zheng20213d}ICCV'21			   &243  &30.0 &33.6 &29.9 &31.0 &30.2 &33.3 &34.8 &31.4 &37.8 &38.6 &31.7 &31.5 &29.0 &23.3 &23.1 &31.3 \\
			MixSTE~\cite{zhang2022mixste}CVPR'22 		  	   & 243 &21.6  &22.0  &20.4  &21.0  &20.8  &24.3  &24.7  &21.9  &26.9  &24.9  &21.2  &21.5  &20.8  &14.7  &15.7  &21.6 \\
			MHFormer~\cite{li2022mhformer}CVPR'22			  &351  &27.7 &32.1 &29.1 &28.9 &30.0 &33.9 &33.0 &31.2 &37.0 &39.3 &30.0 &31.0 &29.4 &22.2 &23.0 &30.5 \\
			Xue~\textit{et al.}~\cite{xue2022boosting}TIP'22   & 243 &25.8  &25.2  &23.3  &23.5  &24.0  &27.4  &27.9  &24.4  &29.3  &30.1  &24.9  &24.1  &23.3  &18.6  &19.7  &24.7 \\
			StridedFormer~\cite{li2022exploiting}TMM'23        & 243 &27.1  &29.4  &26.5  &27.1  &28.6  &33.0  &30.7  &26.8  &38.2  &34.7  &29.1  &29.8  &26.8  &19.1  &19.8  &28.5 \\
			GLA-GCN~\cite{yu2023gla}ICCV'23 				   &243  &20.1 &21.2 &20.0 &19.6 &21.5 &26.7 &23.3 &19.8 &27.0 &29.4 &20.8 &20.1 &19.2 &12.8 &13.8 &21.0 \\
			SCTFormer~\cite{tang20233d}CVPR'23				   & 243 &20.8 	&21.8  &20.0 &20.6 &23.4 &25.0 &23.6 &19.3 &27.8 &26.1 &21.6 &20.6 &19.5 &14.3 &15.1 &21.3 \\
			MotionBERT~\cite{zhu2023motionbert}ICCV'23  & 243 &16.7  &19.9  &17.1 &16.5 &17.4 &18.8 &19.3 &20.5 &24.0 &22.1 &18.6 &\underline{16.8} &16.7 &10.8 &11.5 &17.9 \\
			MotionBERT~\cite{zhu2023motionbert}(finetune)ICCV'23  & 243 &\underline{15.9} &\underline{17.3} &\underline{16.9} &\underline{14.6} &\underline{16.8} &\underline{18.6} &\underline{18.6} &18.4 &\underline{22.0} &\underline{21.8} &\underline{17.3} &16.9 &\underline{16.1} &\underline{10.5} &\underline{11.4} &\underline{16.9} \\
			Zhong~\textit{et al.}~\cite{zhong2023frame}TMM'24  & 243 &20.6  &19.9  &19.5  &19.3  &19.0  &21.7 &21.9  &20.7  &23.5  &24.3  &19.9  &18.5  &19.5  &13.1 &13.9 &19.7 \\
			KTPformer~\cite{peng2024ktpformer}CVPR'24		  				& 243 &19.6 &18.6 &18.5 &18.1 &18.7 &22.1  &20.8 &\underline{18.3} &22.8 &22.4 &18.8 &18.1 &18.4 &13.9  &15.2  &19.0 \\
			MotionAGFormer~\cite{mehraban2024motionagformer}WACV'24 	&243 &-&-&-&-&-&-&-&-&-&-&-&-&-&-&-&19.4 \\
			APP~\cite{zhang2024app}ACMMM'24                                &243 &18.2 &20.6 &18.4 &17.9 &19.5 &21.3 &20.7 &20.6 &25.2 &25.7 &19.3 &18.2 &17.4 &11.3 &12.1 &19.1 \\
			\hline
			\rowcolor{gray!20}Ours         						  &243  &\textbf{14.1} &\textbf{15.8} &\textbf{16.1} &\textbf{14.4} &\textbf{15.8} &\textbf{16.6} &\textbf{16.4} &\textbf{16.6} &\textbf{20.9} &\textbf{20.6} &\textbf{16.0} &\textbf{14.5} &\textbf{14.9} &\textbf{9.0} &\textbf{9.8} &\textbf{15.4}  \\
			\hline
		\end{tabular}
	}
	\label{tab:human3.6mgt}
\end{table*}

\begin{table}[ht]\LARGE
	\caption{Comparison of computational cost and inference speed among different models.}
	\centering
	\resizebox{1\linewidth}{!}{
		\begin{tabular}{l|cccc}
			\hline
			Method                                          		& Param(M) &MACs(G)$\downarrow$  &FPS$\uparrow$ &MPJPE(mm)$\downarrow$	 \\
			\hline
			Poseformer~\cite{zheng20213d} 							&9.5   &0.8      &327    &44.3 \\
			P-STMO~\cite{shan2022p}  								&6.7   &1.0      &148    &42.8 \\
			MixSTE~\cite{zhang2022mixste} 			    			&33.8 & 147.6   &9811  & 40.9 \\
			MotionBERT-scratch~\cite{zhu2023motionbert} 			&42.3 & 174.7   &8021  & 39.2 \\
			MotionAGFormer-B~\cite{mehraban2024motionagformer} 		&11.7  & 64.8    &6511  & 38.4 \\
			MotionAGFormer-L~\cite{mehraban2024motionagformer} 		&19.0  & 105.1   &4026  & 38.4 \\
			\hline
			\rowcolor{gray!20}Ours								    &42.6  & 58.9	 &16231 & 37.7 \\
			\hline
		\end{tabular}%
	}
	\label{tab:macscompare}
\end{table}

\textbf{Results on Human3.6M.} We benchmark our method against other SOTA approaches on the Human3.6M dataset. To ensure a fair comparison, we use 243-frame 2D pose sequences as input. The results are summarized in Table \ref{tab:human3.6m-comparison-action}. For Protocol 1, our method achieves SOTA performance, with an error of 37.7mm. Compared to MotionBERT\cite{zhu2023motionbert}, our method demonstrates a notable improvement, reducing MPJPE from 39.2mm to 37.7mm. It attains the best results in 13 actions, including Direction, Discussion, and Greeting. These findings indicate that our method is more robust across a wide range of actions and generates highly accurate predictions. For Protocol 2, our method continues to demonstrate competitive accuracy, particularly in actions such as Wait and Greeting. To assess the performance upper bound of our model, we use ground-truth 2D joint positions as input, as shown in Table \ref{tab:human3.6mgt}. Our method demonstrates outstanding performance, achieving 15.4mm MPJPE and attaining state-of-the-art accuracy across all actions. This even surpasses MotionBERT, which is fine-tuned on ground truth input (16.9mm vs. 15.4mm). Results indicates that when the model is provided with higher-quality 2D pose detections, it can further enhance its precision.

The comparison of model inference efficiency is presented in Table \ref{tab:macscompare}. All models are evaluated on the same system equipped with an RTX 3090 GPU to ensure fair comparisons. Compared to sequence-to-frame models, such as PoseFormer and P-STMO, our method achieves both higher accuracy and faster inference speed. For sequence-to-sequence prediction models with larger parameter sizes, such as MotioBERT and MixSTE, our method benefits from SCT, which reduces the computational complexity of the attention mechanism. As a result, our model achieves a higher FPS, enabling it to process more temporal frames during inference. Furthermore, our method maintains state-of-the-art accuracy despite the efficiency gains.

\begin{table}[ht]
	\centering
	\caption{Quantitative Results on MPI-INF-3DHP. $f$ denotes the number of input frames in the 3DHP dataset. The best results are highlighted in bold.}
	\resizebox{1\linewidth}{!}{
		\begin{tabular}{lc | ccc}
			\hline
			Model           &$f$ & PCK$\uparrow$ & AUC$\uparrow$ & MPJPE$\downarrow$ \\
			\hline
			P-STMO~\cite{shan2022p}                                   		& 81 & 97.9 & 75.8 & 32.2 \\
			HDformer~\cite{chen2023hdformer}								& 96 & \textbf{98.7} & 72.9 & 37.2 \\
			Einfalt \textit{et al.}~\cite{einfalt2023uplift} 				& 81 & 95.4 & 67.6 & 46.9 \\
			MixSTE~\cite{zhang2022mixste}                                   & 27 & 94.4 & 66.5 & 54.9 \\
			STCFormer~\cite{tang20233d}                            			& 81 & \textbf{98.7} & 83.9 & 23.1 \\
			PoseFormerV2~\cite{zhao2023poseformerv2}                      	& 81 & 97.9 & 78.8 & 27.8 \\
			GLA-GCN~\cite{yu2023gla}                           				& 81 & 98.5 & 79.1 & 27.7 \\
			MotionAGFormer-B~\cite{mehraban2024motionagformer}              & 81 & 98.3 & 84.2 & 18.2 \\
			MotionAGFormer-L~\cite{mehraban2024motionagformer}				& 81 & 98.2 & 85.3 & 16.2 \\
			FTCM-L~\cite{tang2023ftcm}										& 81 & 98.0	& 79.8 & 31.2 \\
			\hline   
			\rowcolor{gray!20}Ours             								& 81 & \textbf{98.7} & \textbf{86.6} & \textbf{16.0} \\
			\hline
		\end{tabular}
	}
	\label{tab:table_3dhp}
\end{table}
\textbf{Results on MPI-INF-3DHP.} Table \ref{tab:table_3dhp} summarizes the performance of our method on the MPI-INF-3DHP dataset. Using ground-truth 2D poses as input, our method achieves state-of-the-art results across all three metrics: PCK, AUC, and MPJPE. Notably, we observe a substantial reduction in MPJPE compared to HDFormer and STCFormer (from 37.2mm / 23.1mm to 16.0mm), despite both methods achieving similar PCK scores. Compared to MotionAGFormer-L, our method surpasses it by 0.5\% in PCK and 0.2mm in MPJPE. Furthermore, with the integration of SCT, our method exhibits lower computational complexity. Since the 3DHP dataset contains both indoor and outdoor environments, these results highlight the strong generalization ability and robustness of our method across diverse scenarios.

\subsection{Qualitative Analysis}
\begin{figure}[ht]
	\centering
	\includegraphics[width=\linewidth]{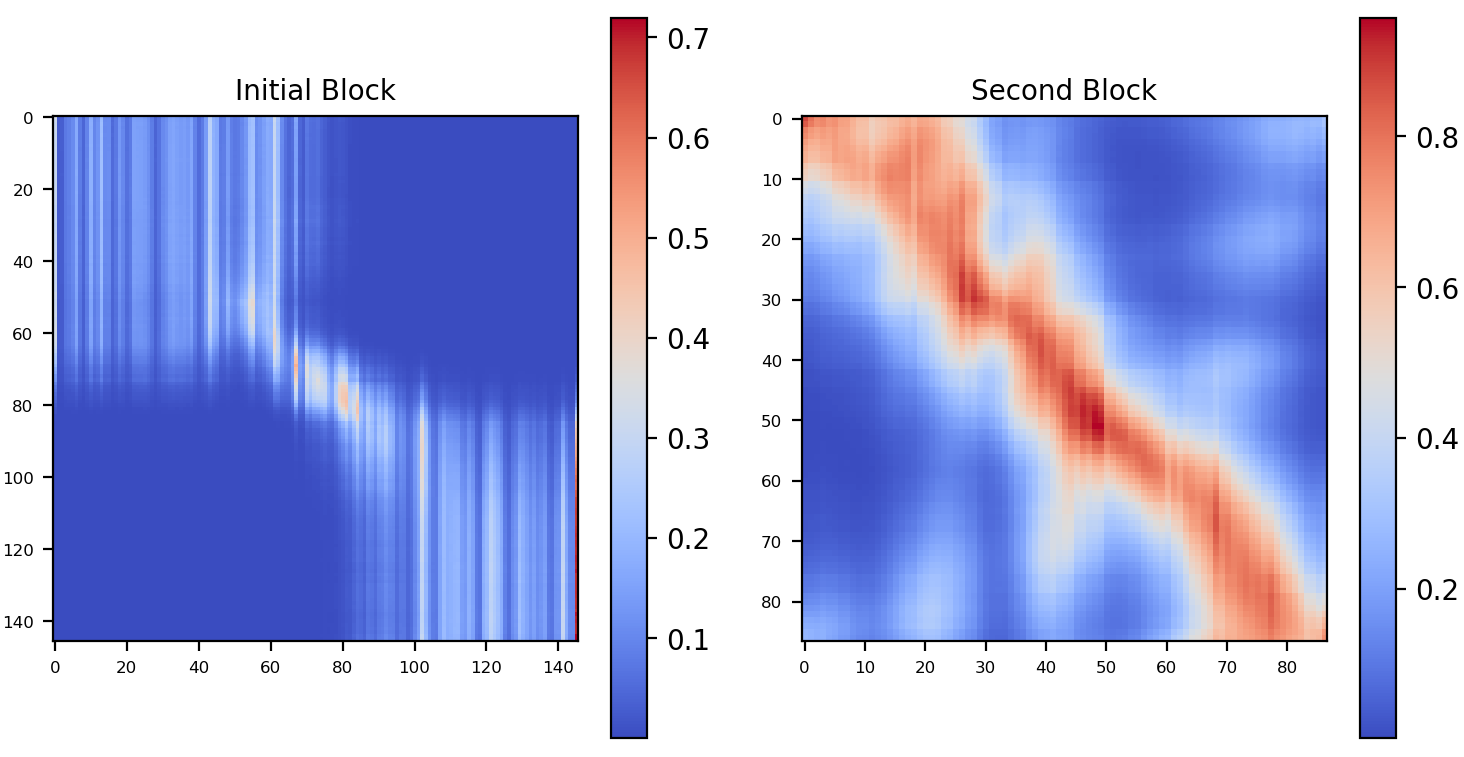}
	\caption{Temporal attention map for the initial block and the second block. Darker colors between frames indicate stronger correlations between the frames.}
	\label{fig:heatmap}
\end{figure}
\begin{figure}[ht]
	\centering
	\includegraphics[width=\linewidth]{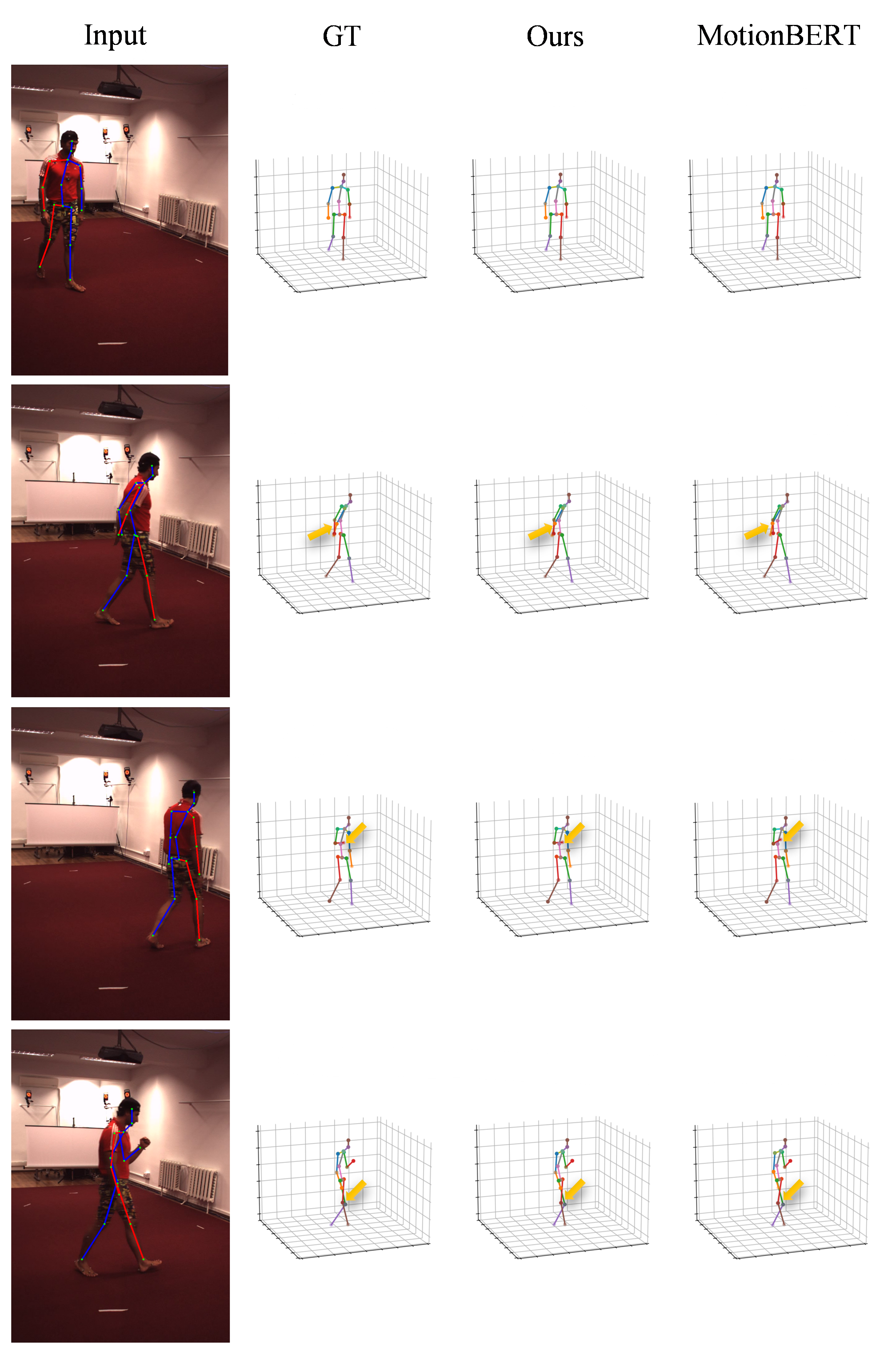}
	\caption{A visual comparison between the improved MotionBERT method and the original version is presented for the Walking of the S9 subject in the Human3.6M dataset. The differences in predicted joint positions are highlighted in the figure, showing that the improved method yields predictions that are closer to the ground truth.}
	\label{fig:keshihua}
\end{figure}
\begin{figure*}[ht]
	\centering
	\includegraphics[width=\linewidth]{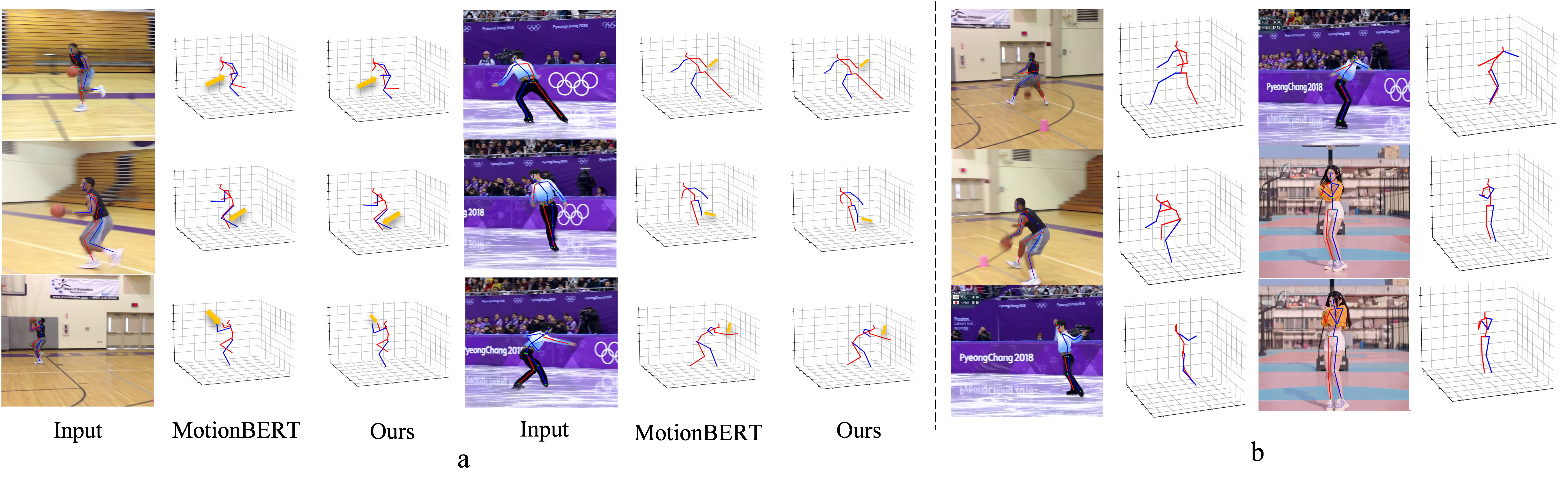}
	\caption{The visualization results on in-the-wild videos are shown. Subgraph a presents a comparison on various sports actions, while Subgraph b demonstrates that even in cases of motion blur or significant deviations in 2D pose predictions, our method is still capable of reconstructing relatively accurate 3D poses.}
	\label{fig:inthewild_keshihua}
\end{figure*}

Our SCT module is primarily designed to reduce redundancy in the temporal sequence. To observe the dependencies between different time frames, we generated a temporal attention heatmap. Figure \ref{fig:heatmap} presents the attention map of the temporal transformer block for the Direction of subject S9 in the test set. In the initial block, the regions with higher weights are distributed roughly near the diagonal but are relatively scattered. After the hidden features pass through the second SCT-integrated block, the temporal sequence length is compressed. Higher-weight values cluster along the diagonal, indicating strong dependencies between adjacent frames in the compressed sequence.

Figure \ref{fig:keshihua} qualitatively compares our method with MotionBERT. In the absence of occlusion, we observe that both our method and MotionBERT can accurately reconstruct the skeleton. When the subject performs lateral movements, the opposite side of the body, such as the left arm and left leg, experiences significant occlusion. However, our method reconstructs the occluded limbs with greater accuracy, bringing them much closer to the ground truth. This demonstrates that our method is more effective in handling postures with severe occlusions.

Figure \ref{fig:inthewild_keshihua}-a show the qualitative comparison between our method, MotionBERT on in-the-wild videos. In some hand and leg movements, our method can more effectively capture local details. Figure \ref{fig:inthewild_keshihua}-b shows the qualitative results on challenging in-the-wild videos. The results demonstrate that our method can produce precise 3D pose estimation.

\subsection{Ablation Studies}
A series of ablation studies are conducted on the Human3.6M dataset to explore different module design choices.
\subsubsection{Network Architecture}
\begin{table}[ht]
	\caption{We compare model parameters, computational complexity, and error by varying the number of network layers, the feature dimension of the hidden layer, and the compression factor. Here, $L$ denotes the network depth, $C$ represents the hidden layer feature dimension, and $\sigma$ indicates the compression factor.}
	\label{tab:ablation_netartic}
	\centering
	\resizebox{1\linewidth}{!}{
		\begin{tabular}{@{}lcc | ccc@{}}
			\hline
			$L$		  			&$C$    		& $\sigma$		  & Params(M)  &MACs(G)  &MPJPE(mm)	\\
			\hline
			4	&256  &0.6  &12.8    &21.2   &39.6  \\
			5   &256  &0.5  &16.0	 &16.9   &39.3  \\
			5   &256  &0.6  &16.0	 &22.4   &39.1  \\
			5   &256  &0.7  &16.0	 &29.9   &39.2  \\
			6   &256  &0.5  &19.2    &17.2   &39.2   \\
			6   &256  &0.6  &19.2    &23.2   &38.9   \\
			6   &256  &0.7  &19.2    &32.2   &39.0   \\
			7   &256  &0.6  &22.3    &23.7   &39.1   \\
			\hline
			4	&512  &0.6  &34.1    &55.6   &39.3  \\
			4	&512  &0.7  &34.1    &71.9   &39.1  \\
			5	&512  &0.5  &42.6    &44.4   &38.3  \\
			5	&512  &0.6  &42.6    &58.9   &37.7  \\
			5	&512  &0.7  &42.6 	 &78.6   &38.2  \\
			6	&512  &0.5  &50.8    &45.1   &38.1  \\
			6	&512  &0.6  &50.8    &60.9   &37.9  \\
			6	&512  &0.7  &50.8 	 &83.4   &38.2  \\
			\hline
		\end{tabular}
	}
\end{table}
We conduct ablation experiments to investigate the impact of different network depths and feature dimensions, as shown in Table \ref{tab:ablation_netartic}. When the feature dimension is set to 256, the model has fewer parameters and lower computational complexity but exhibits a relatively higher overall error. We also observe that reducing the compression coefficient decreases the computational load but results in reduced accuracy. When the feature dimension increases to 512, the overall error significantly decreases. Furthermore, we find that for model depths of 5 and 6, the error reaches its minimum when $\sigma$ is set to 0.6. However, when the model depth deviates from 5—either increasing or decreasing—the error tends to rise. The model achieves a favorable balance between accuracy and computational efficiency when using 5 layers and a feature dimension of 512.

\subsubsection{The effectiveness of the proposed module}
\begin{table}[ht]\LARGE
	\caption{Ablation study for each module employed in our approach, evaluated on Human3.6M with MPJPE(mm) and MACs(G) metrics. Vanilla Transformer denotes a network configuration that includes only standard Transformer blocks, without incorporating any additional components. FD-loss indicates whether the frequency-domain loss function is applied during training.}
	\label{tab:ablation_per_module}
	\centering
	\resizebox{1\linewidth}{!}{
		\begin{tabular}{@{}lccc | cc@{}}
			\hline
			Module		  			&LPG    		& SCT		  & FD-loss   	&MPJPE(mm)      & MACs(G)	\\
			\hline
			Vanilla Transformer	    & -             & -           & -     	 	& 39.2     & 174.7	 \\
			\checkmark    			& \checkmark    &             &            	& 38.7     & 174.9   \\
			\checkmark    			& \checkmark    & \checkmark  &            	& 38.0     & 58.9    \\
			\checkmark   			& \checkmark    & 			  & \checkmark  & 39.0     & 174.9   \\
			\checkmark   			& \checkmark    & \checkmark  & \checkmark  & 37.7     & 58.9    \\
			\hline
		\end{tabular}
	}
\end{table}

\begin{figure}[ht]
	\centering
	\includegraphics[width=\linewidth]{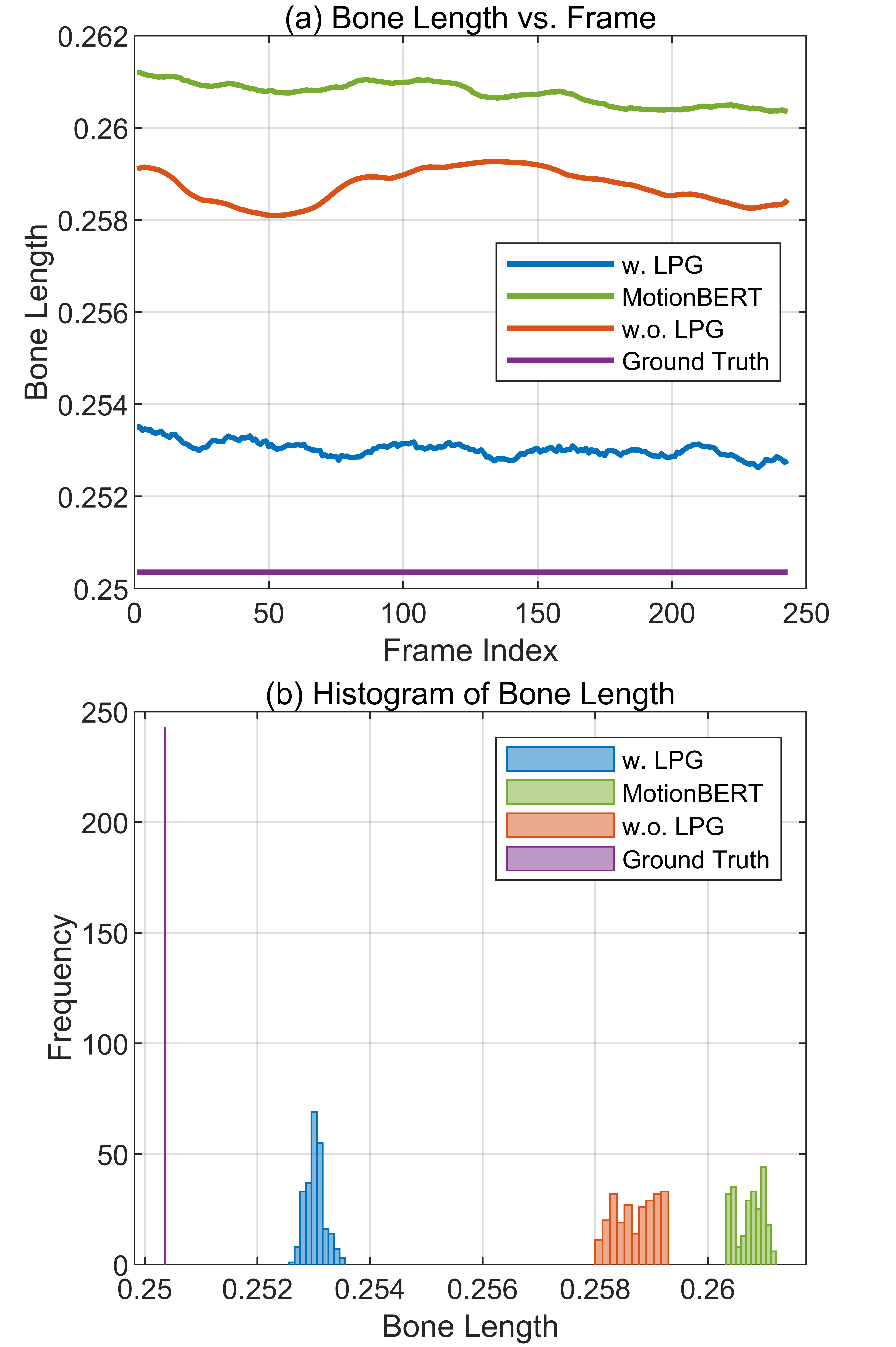}
	\caption{The bone length and frequency distribution before and after adding LPG. Here, we compute the mean bone length from the left elbow to the left hand across different action videos of subject S9. In subfigure a, smaller horizontal fluctuations indicate greater stability in bone length. In subfigure b, a more concentrated frequency distribution corresponds to more stable predicted bone lengths.}
	\label{fig:bonecompare}
\end{figure}

To verify the effectiveness of the proposed modules, we conduct ablation experiments by progressively integrating each module into the model, as shown in Table \ref{tab:ablation_per_module}. When our model consists of only standard Transformer blocks, the error and computational cost are 39.2mm and 174.7G, respectively. Adding the LPG module reduces the error by 0.5mm. Although LPG increases MACs by 0.2G, this overhead is negligible relative to the overall computational cost. We also compare the variance in the distribution of bone lengths in consecutive pose sequences before and after incorporating the LPG module. In a continuous sequence of frames, the bone lengths of the same subject remain consistent. We compute the average variance of bone lengths across all frames for subject S9 in the Human3.6M dataset. The variances before adding LPG, after adding LPG, and the ground truth are $1.2\times 10^{-7}$, $7.4 \times 10^{-8}$, $6.7\times 10^{-15}$ respectively, indicating that the method with LPG significantly reduces the variance. Taking the bone segment from the left elbow to the left hand as an example, Figure \ref{fig:bonecompare} (a) and (b) demonstrate that the LPG module yields a bone length more consistent with the ground truth while also exhibiting a more stable distribution.

Building upon LPG, we replace the original temporal sequence encoder with SCT, leading to a 0.7mm reduction in error and a computational cost reduction of 116G, effectively reducing the original computational load. This demonstrates that the SCT module can effectively compress sequences while enabling the model to fully learn compressed sequence features. When only the frequency-domain loss (FD-loss) is applied during training, the reduction in error is minimal. However, when SCT and FD-loss are applied together, the error further decreases by 0.3mm. This improvement occurs because FD-loss supervises the model by aligning the power spectra of predicted and ground-truth poses, thereby facilitating both sequence compression and spectral feature utilization.

\subsubsection{Effectiveness of High-Frequency Information and Analysis of Frequency Domain Loss Weight}
\begin{table}[ht]\small
	\caption{(Top) The influence of high-frequency information on model error and computational cost. (Bottom) The impact of different frequency-domain loss weights on error.}
	\label{tab:ablation_sigma}
	\centering
		\begin{tabular}{@{}c | cc@{}}
			\hline
			$\sigma$	  	&MPJPE(mm)  & MACs(G)	  \\
			\hline
			0.3				& 39.6    & 25.2  \\
			0.4			    & 39.0    & 33.7  \\
			0.5    			& 38.3    & 44.4  \\
			0.6   			& 37.7    & 58.9  \\
			0.7 			& 38.2    & 78.6  \\
			0.8			    & 38.6    & 104.4  \\
			0.9             & 38.8    & 140.1  \\
			\hline
			$\lambda$	  	&MPJPE(mm)  & P-MPJPE(mm) \\
			\hline
			0	&38.0 & 32.0 \\
			0.5	&37.9 & 31.9 \\
			1	&37.7 & 31.7  \\
			2	&37.9 & 31.8  \\
			5	&37.9 & 31.9 \\
			10	&38.1 & 32.0 \\		
			\hline
		\end{tabular}
\end{table}
In the SCT module, the compression coefficient $\sigma$ is a key hyperparameter that regulates how much sequence information is retained and the model's computational efficiency. As shown in Table \ref{tab:ablation_sigma} top, we evaluate $\sigma$ values ranging from 0.3 to 0.9 in increments of 0.1. We observed that when $\sigma$ was set to 0.3, the model filtered out a substantial amount of high-frequency information, leading to significant reconstruction errors due to the loss of sequence details. As $\sigma$ increased from 0.3 to 0.5, the retention of fine-grained details improved, resulting in a gradual reduction in error, albeit at the cost of increased computational complexity. At $\sigma=0.6$, a balance between accuracy and computational cost is achieved, yielding the lowest error while maintaining relatively low computational cost. As additional high-frequency details were preserved, the error continued to increase. These high-frequency components can be considered as noise, and their removal not only enhances computational efficiency but also improves the accuracy of the results. Therefore, setting $\sigma$ to 0.6 provides a well-balanced trade-off.

We also examined the effect of the frequency domain loss weight (see in Table \ref{tab:ablation_sigma} bottom). As the loss weight increased, the model became overly constrained by frequency domain information, which impaired the accurate reconstruction of the 3D pose sequence. The optimal results were obtained when $\lambda$ was set to 1.

\begin{table}[ht]\LARGE
	\caption{Quantitative Comparison of Various 3D Human Pose Estimators Trained with and without SCT on the Human3.6M Dataset. FD-loss indicates the frequency-domain loss function.}
	\label{tab:ablation_adapative}
	\centering
	\resizebox{1\linewidth}{!}{
		\begin{tabular}{@{}l | ccc@{}}
			\hline
			Method	  		&MPJPE(mm)  &Params(M)  & MACs(G)	  \\
			\hline
			MixSTE~\cite{zhang2022mixste}			& 40.9    & 33.78   & 138.6    \\ 
			+SCT+FD-loss    						& 40.1(-0.8)    & 33.78   & 39.5(-99.1)	\\
			\hline
			MotionAGFormer~\cite{mehraban2024motionagformer}  & 38.5  		  & 11.72    & 64.8  \\
			+SCT+FD-loss    								  & 38.2(-0.3)    & 11.72    	 & 12.1(-52.7)  \\
			\hline
			MotionBERT-lite(scratch)~\cite{zhu2023motionbert} 	& 39.8  		&16.0 	& 71.3     \\
			+SCT+FD-loss									& 38.8(-1.0)    &16.0  & 29.9(-41.4)   \\
			\hline
			MotionBERT-lite(finetune)~\cite{zhu2023motionbert} 	& 37.9  		&16.0 	& 71.3     \\
			+SCT+FD-loss									& 36.7(-1.2)		&16.0   & 29.9(-41.4)   \\
			\hline
			RePose~\cite{sun2024repose}						& 36.5		&16.0  & 71.3	  \\
			APP~\cite{zhang2024app}							& 37.0		&18.2  &211.8		\\
			\hline
		\end{tabular}
	}
\end{table}
To evaluate SCT’s ability to reduce redundancy in pose sequences, we additionally selected three other mainstream transformer-based 3D pose estimators for comparison. Referring to the model structure in this paper, we replaced the first five transformer encoder blocks used for processing time series in the original model with SCT. For each method, $\sigma$ was set to 0.6 and $\lambda$ was set to 1 during training. However, fine-tuning $\sigma$ for each model individually may lead to even better results. As shown in Table \ref{tab:ablation_adapative}, our SCT module effectively reduces computational cost in most methods while also providing modest accuracy improvements. Specifically, compared to other MotionBERT-lite-based methods, such as RePose \cite{sun2024repose} and APP \cite{zhang2024app}, our method achieves the second-highest accuracy (36.5mm vs. 36.7mm) while demonstrating a significant advantage in computational efficiency (29.9G vs. 71.3G).

\subsubsection{Sequence Recovery Method}
\begin{table}[ht]\LARGE
	\caption{Impact of Different Upsampling Locations and Interpolation Methods on MPJPE and Computational Cost.}
	\label{tab:ablation_upsampling}
	\centering
	\resizebox{1\linewidth}{!}{
		\begin{tabular}{@{}cc | cc@{}}
			\hline
			Restoration Position 			& Upsampling Method  	& MACs(G) &  MPJPE(mm)	\\
			\hline
			Final Layer  					& Nearest Interpolation       	  		&58.9      & 38.4   \\
			Final Layer						& Cross Attention						&61.4	   & 37.9 	\\
			All Layers 						& Nearest Interpolation            		&58.9      & 37.8   \\
			All Layers 						& Linear Interpolation             		&58.9      & 37.7   \\
			\hline
		\end{tabular}
	}
\end{table}

Restoring full-length posture sequences involves an up-sampling problem concerning temporal frames, where interpolation methods are commonly used non-parametric methods in computer vision. We explore two restoration methods: recovering only the output of the final layer and restoring the hidden feature sequences of all layers. As shown in Table \ref{tab:ablation_upsampling}, we evaluate two restoration methods for the final layer. Nearest-neighbor interpolation introduces no additional computational cost but results in higher error, as the final layer’s feature sequence, after multiple compression stages, lacks sufficient details to retain complete sequence information. We also apply the cross-attention-based restoration method proposed in HoT \cite{li2024hourglass} to recover temporal resolution, but this approach increases computational cost by 2.5G. For restoration across all layers, linear interpolation outperforms nearest-neighbor interpolation, as all layers contain feature information at different temporal resolutions, and linear interpolation helps generate smoother pose sequences. Compared to cross-attention-based restoration, our interpolation-based upsampling introduces minimal additional computational overhead while achieving higher accuracy.

\section{Conclusion}
This paper introduces the Spectral Compression Transformer and Line Pose Graph to enhance the efficiency and accuracy of 3D human pose estimation in long video sequences captured from a monocular camera. Our method reveals that models processing long video sequences contain redundant information in their hidden features. By leveraging spectral compression, we improve computational efficiency while also enhancing model accuracy. Additionally, the LPG module effectively incorporates input 2D pose priors, enabling the integration of complementary information between bone positions and joint positions. We achieve competitive performance on two widely used datasets, Human3.6M and MPI-INF-3DHP, while significantly reducing computational cost. In addition, the proposed method also exhibits compatibility with other 3D human pose estimation approaches.

\textbf{Future Work} In SCT, the compression coefficient $\sigma$ is a critical hyperparameter that must be hand-tuned, which limits the performance gains when applying the SCT encoder across different methods. Although the paper demonstrates that $\sigma=0.6$ produces satisfactory results in both our network and other approaches, this finding lacks theoretical validation and extensive experimental evidence. We therefore aim to automatically determine the optimal value of $\sigma$ in various models, thereby achieving a balance between computational efficiency and accuracy. Additionally, we plan to further optimize the upsampling method and embedding capability to enhance the model's overall performance.

\appendix
\section{Appendix}
\label{Appendix}
\begin{figure*}[ht]
	\centering
	\includegraphics[width=\linewidth]{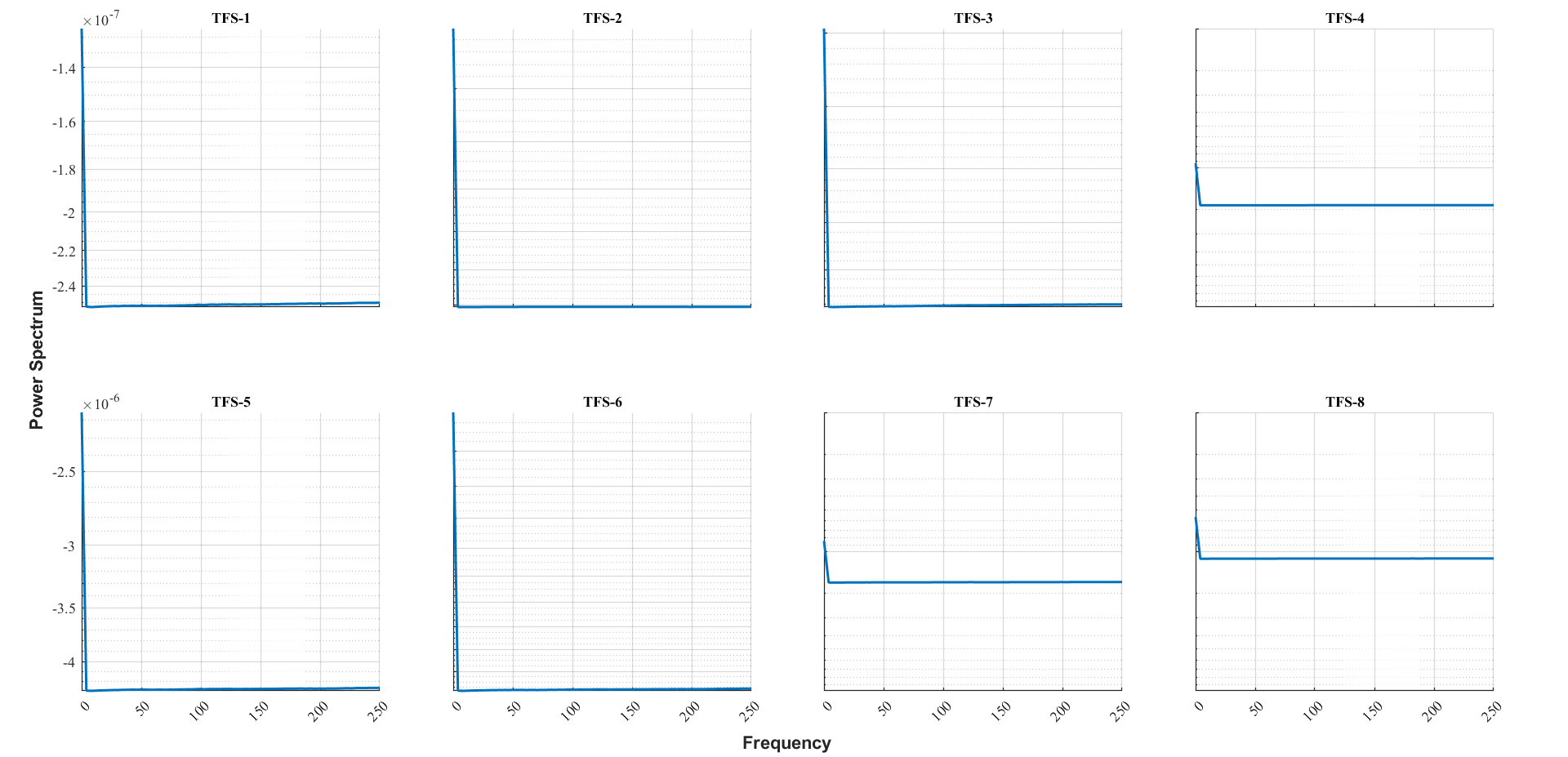}
	\caption{The power spectrum of each TFS in MixSTE. The horizontal axis represents the frequency range, while the vertical axis represents the power spectrum.}
	\label{fig:supp_TFSinfreq}
\end{figure*}
As shown in Figure \ref{fig:supp_TFSinfreq}, the hidden features of each block in MixSTE are transformed into TFS and then mapped to the frequency domain. We observe that the energy of TFS is primarily concentrated in the low-frequency region, whereas the high-frequency components exhibit an extended tail. This phenomenon is consistently observed across hidden features of each inter-block.

\bibliographystyle{IEEEtran}
\bibliography{references.bib}

\begin{thebibliography}{10}
\providecommand{\url}[1]{#1}
\csname url@samestyle\endcsname
\providecommand{\newblock}{\relax}
\providecommand{\bibinfo}[2]{#2}
\providecommand{\BIBentrySTDinterwordspacing}{\spaceskip=0pt\relax}
\providecommand{\BIBentryALTinterwordstretchfactor}{4}
\providecommand{\BIBentryALTinterwordspacing}{\spaceskip=\fontdimen2\font plus
\BIBentryALTinterwordstretchfactor\fontdimen3\font minus
  \fontdimen4\font\relax}
\providecommand{\BIBforeignlanguage}[2]{{%
\expandafter\ifx\csname l@#1\endcsname\relax
\typeout{** WARNING: IEEEtran.bst: No hyphenation pattern has been}%
\typeout{** loaded for the language `#1'. Using the pattern for}%
\typeout{** the default language instead.}%
\else
\language=\csname l@#1\endcsname
\fi
#2}}
\providecommand{\BIBdecl}{\relax}
\BIBdecl

\bibitem{luvizon2020multi}
D.~C. Luvizon, D.~Picard, and H.~Tabia, ``Multi-task deep learning for
  real-time 3d human pose estimation and action recognition,'' \emph{IEEE
  transactions on pattern analysis and machine intelligence}, vol.~43, no.~8,
  pp. 2752--2764, 2020.

\bibitem{munea2020progress}
T.~L. Munea, Y.~Z. Jembre, H.~T. Weldegebriel, L.~Chen, C.~Huang, and C.~Yang,
  ``The progress of human pose estimation: A survey and taxonomy of models
  applied in 2d human pose estimation,'' \emph{IEEE Access}, vol.~8, pp.
  133\,330--133\,348, 2020.

\bibitem{wiederer2020traffic}
J.~Wiederer, A.~Bouazizi, U.~Kressel, and V.~Belagiannis, ``Traffic control
  gesture recognition for autonomous vehicles,'' in \emph{2020 IEEE/RSJ
  International Conference on Intelligent Robots and Systems (IROS)}.\hskip 1em
  plus 0.5em minus 0.4em\relax IEEE, 2020, pp. 10\,676--10\,683.

\bibitem{cai2019exploiting}
Y.~Cai, L.~Ge, J.~Liu, J.~Cai, T.-J. Cham, J.~Yuan, and N.~M. Thalmann,
  ``Exploiting spatial-temporal relationships for 3d pose estimation via graph
  convolutional networks,'' in \emph{Proceedings of the IEEE/CVF international
  conference on computer vision}, 2019, pp. 2272--2281.

\bibitem{chen2021anatomy}
T.~Chen, C.~Fang, X.~Shen, Y.~Zhu, Z.~Chen, and J.~Luo, ``Anatomy-aware 3d
  human pose estimation with bone-based pose decomposition,'' \emph{IEEE
  Transactions on Circuits and Systems for Video Technology}, vol.~32, no.~1,
  pp. 198--209, 2021.

\bibitem{wang2020motion}
J.~Wang, S.~Yan, Y.~Xiong, and D.~Lin, ``Motion guided 3d pose estimation from
  videos,'' in \emph{European conference on computer vision}.\hskip 1em plus
  0.5em minus 0.4em\relax Springer, 2020, pp. 764--780.

\bibitem{shan2023diffusion}
W.~Shan, Z.~Liu, X.~Zhang, Z.~Wang, K.~Han, S.~Wang, S.~Ma, and W.~Gao,
  ``Diffusion-based 3d human pose estimation with multi-hypothesis
  aggregation,'' in \emph{Proceedings of the IEEE/CVF International Conference
  on Computer Vision}, 2023, pp. 14\,761--14\,771.

\bibitem{foo2023unified}
L.~G. Foo, T.~Li, H.~Rahmani, Q.~Ke, and J.~Liu, ``Unified pose sequence
  modeling,'' in \emph{Proceedings of the IEEE/CVF Conference on Computer
  Vision and Pattern Recognition}, 2023, pp. 13\,019--13\,030.

\bibitem{li2021tokenpose}
Y.~Li, S.~Zhang, Z.~Wang, S.~Yang, W.~Yang, S.-T. Xia, and E.~Zhou,
  ``Tokenpose: Learning keypoint tokens for human pose estimation,'' in
  \emph{Proceedings of the IEEE/CVF International conference on computer
  vision}, 2021, pp. 11\,313--11\,322.

\bibitem{chen2023hdformer}
H.~Chen, J.-Y. He, W.~Xiang, Z.-Q. Cheng, W.~Liu, H.~Liu, B.~Luo, Y.~Geng, and
  X.~Xie, ``Hdformer: High-order directed transformer for 3d human pose
  estimation,'' \emph{arXiv preprint arXiv:2302.01825}, 2023.

\bibitem{zhu2023motionbert}
W.~Zhu, X.~Ma, Z.~Liu, L.~Liu, W.~Wu, and Y.~Wang, ``Motionbert: A unified
  perspective on learning human motion representations,'' in \emph{Proceedings
  of the IEEE/CVF International Conference on Computer Vision}, 2023, pp.
  15\,085--15\,099.

\bibitem{wang2024utilizing}
Y.~Wang, M.~Li, and H.~Yan, ``Utilizing motion segmentation for optimizing the
  temporal adjacency matrix in 3d human pose estimation,''
  \emph{Neurocomputing}, vol. 600, p. 128153, 2024.

\bibitem{xu2020deep}
J.~Xu, Z.~Yu, B.~Ni, J.~Yang, X.~Yang, and W.~Zhang, ``Deep kinematics analysis
  for monocular 3d human pose estimation,'' in \emph{Proceedings of the
  IEEE/CVF Conference on computer vision and Pattern recognition}, 2020, pp.
  899--908.

\bibitem{li2020cascaded}
S.~Li, L.~Ke, K.~Pratama, Y.-W. Tai, C.-K. Tang, and K.-T. Cheng, ``Cascaded
  deep monocular 3d human pose estimation with evolutionary training data,'' in
  \emph{Proceedings of the IEEE/CVF conference on computer vision and pattern
  recognition}, 2020, pp. 6173--6183.

\bibitem{peng2024ktpformer}
J.~Peng, Y.~Zhou, and P.~Mok, ``Ktpformer: Kinematics and trajectory prior
  knowledge-enhanced transformer for 3d human pose estimation,'' in
  \emph{Proceedings of the IEEE/CVF Conference on Computer Vision and Pattern
  Recognition}, 2024, pp. 1123--1132.

\bibitem{zhao2023poseformerv2}
Q.~Zhao, C.~Zheng, M.~Liu, P.~Wang, and C.~Chen, ``Poseformerv2: Exploring
  frequency domain for efficient and robust 3d human pose estimation,'' in
  \emph{Proceedings of the IEEE/CVF Conference on Computer Vision and Pattern
  Recognition}, 2023, pp. 8877--8886.

\bibitem{li2024hourglass}
W.~Li, M.~Liu, H.~Liu, P.~Wang, J.~Cai, and N.~Sebe, ``Hourglass tokenizer for
  efficient transformer-based 3d human pose estimation,'' in \emph{Proceedings
  of the IEEE/CVF Conference on Computer Vision and Pattern Recognition}, 2024,
  pp. 604--613.

\bibitem{ionescu2013human3}
C.~Ionescu, D.~Papava, V.~Olaru, and C.~Sminchisescu, ``Human3. 6m: Large scale
  datasets and predictive methods for 3d human sensing in natural
  environments,'' \emph{IEEE transactions on pattern analysis and machine
  intelligence}, vol.~36, no.~7, pp. 1325--1339, 2013.

\bibitem{mehta2017monocular}
D.~Mehta, H.~Rhodin, D.~Casas, P.~Fua, O.~Sotnychenko, W.~Xu, and C.~Theobalt,
  ``Monocular 3d human pose estimation in the wild using improved cnn
  supervision,'' in \emph{2017 international conference on 3D vision
  (3DV)}.\hskip 1em plus 0.5em minus 0.4em\relax IEEE, 2017, pp. 506--516.

\bibitem{vaswani2017attention}
A.~Vaswani, N.~Shazeer, N.~Parmar, J.~Uszkoreit, L.~Jones, A.~N. Gomez,
  {\L}.~Kaiser, and I.~Polosukhin, ``Attention is all you need,''
  \emph{Advances in neural information processing systems}, vol.~30, 2017.

\bibitem{zheng20213d}
C.~Zheng, S.~Zhu, M.~Mendieta, T.~Yang, C.~Chen, and Z.~Ding, ``3d human pose
  estimation with spatial and temporal transformers,'' in \emph{Proceedings of
  the IEEE/CVF international conference on computer vision}, 2021, pp.
  11\,656--11\,665.

\bibitem{qian2023hstformer}
X.~Qian, Y.~Tang, N.~Zhang, M.~Han, J.~Xiao, M.-C. Huang, and R.-S. Lin,
  ``Hstformer: Hierarchical spatial-temporal transformers for 3d human pose
  estimation,'' \emph{arXiv preprint arXiv:2301.07322}, 2023.

\bibitem{wan2021encoder}
Z.~Wan, Z.~Li, M.~Tian, J.~Liu, S.~Yi, and H.~Li, ``Encoder-decoder with
  multi-level attention for 3d human shape and pose estimation,'' in
  \emph{Proceedings of the IEEE/CVF International Conference on Computer
  Vision}, 2021, pp. 13\,033--13\,042.

\bibitem{liu2020attention}
R.~Liu, J.~Shen, H.~Wang, C.~Chen, S.-c. Cheung, and V.~Asari, ``Attention
  mechanism exploits temporal contexts: Real-time 3d human pose
  reconstruction,'' in \emph{Proceedings of the IEEE/CVF conference on computer
  vision and pattern recognition}, 2020, pp. 5064--5073.

\bibitem{lin2021end}
K.~Lin, L.~Wang, and Z.~Liu, ``End-to-end human pose and mesh reconstruction
  with transformers,'' in \emph{Proceedings of the IEEE/CVF conference on
  computer vision and pattern recognition}, 2021, pp. 1954--1963.

\bibitem{li2022mhformer}
W.~Li, H.~Liu, H.~Tang, P.~Wang, and L.~Van~Gool, ``Mhformer: Multi-hypothesis
  transformer for 3d human pose estimation,'' in \emph{Proceedings of the
  IEEE/CVF Conference on Computer Vision and Pattern Recognition}, 2022, pp.
  13\,147--13\,156.

\bibitem{zhang2022mixste}
J.~Zhang, Z.~Tu, J.~Yang, Y.~Chen, and J.~Yuan, ``Mixste: Seq2seq mixed
  spatio-temporal encoder for 3d human pose estimation in video,'' in
  \emph{Proceedings of the IEEE/CVF conference on computer vision and pattern
  recognition}, 2022, pp. 13\,232--13\,242.

\bibitem{mehraban2024motionagformer}
S.~Mehraban, V.~Adeli, and B.~Taati, ``Motionagformer: Enhancing 3d human pose
  estimation with a transformer-gcnformer network,'' in \emph{Proceedings of
  the IEEE/CVF Winter Conference on Applications of Computer Vision}, 2024, pp.
  6920--6930.

\bibitem{zhou2023dual}
L.~Zhou, Y.~Chen, and J.~Wang, ``Dual-path transformer for 3d human pose
  estimation,'' \emph{IEEE Transactions on Circuits and Systems for Video
  Technology}, 2023.

\bibitem{tang2023ftcm}
Z.~Tang, Y.~Hao, J.~Li, and R.~Hong, ``Ftcm: Frequency-temporal collaborative
  module for efficient 3d human pose estimation in video,'' \emph{IEEE
  Transactions on Circuits and Systems for Video Technology}, vol.~34, no.~2,
  pp. 911--923, 2023.

\bibitem{zeng2022deciwatch}
A.~Zeng, X.~Ju, L.~Yang, R.~Gao, X.~Zhu, B.~Dai, and Q.~Xu, ``Deciwatch: A
  simple baseline for 10$\times$ efficient 2d and 3d pose estimation,'' in
  \emph{European Conference on Computer Vision}.\hskip 1em plus 0.5em minus
  0.4em\relax Springer, 2022, pp. 607--624.

\bibitem{einfalt2023uplift}
M.~Einfalt, K.~Ludwig, and R.~Lienhart, ``Uplift and upsample: Efficient 3d
  human pose estimation with uplifting transformers,'' in \emph{Proceedings of
  the IEEE/CVF Winter Conference on Applications of Computer Vision}, 2023, pp.
  2903--2913.

\bibitem{tang20233d}
Z.~Tang, Z.~Qiu, Y.~Hao, R.~Hong, and T.~Yao, ``3d human pose estimation with
  spatio-temporal criss-cross attention,'' in \emph{Proceedings of the IEEE/CVF
  Conference on Computer Vision and Pattern Recognition}, 2023, pp. 4790--4799.

\bibitem{li2022exploiting}
W.~Li, H.~Liu, R.~Ding, M.~Liu, P.~Wang, and W.~Yang, ``Exploiting temporal
  contexts with strided transformer for 3d human pose estimation,'' \emph{IEEE
  Transactions on Multimedia}, vol.~25, pp. 1282--1293, 2022.

\bibitem{zhang2023learning}
J.~Zhang, K.~Gong, X.~Wang, and J.~Feng, ``Learning to augment poses for 3d
  human pose estimation in images and videos,'' \emph{IEEE Transactions on
  Pattern Analysis and Machine Intelligence}, vol.~45, no.~8, pp.
  10\,012--10\,026, 2023.

\bibitem{wandt2022elepose}
B.~Wandt, J.~J. Little, and H.~Rhodin, ``Elepose: Unsupervised 3d human pose
  estimation by predicting camera elevation and learning normalizing flows on
  2d poses,'' in \emph{Proceedings of the IEEE/CVF Conference on Computer
  Vision and Pattern Recognition}, 2022, pp. 6635--6645.

\bibitem{qin2021fcanet}
Z.~Qin, P.~Zhang, F.~Wu, and X.~Li, ``Fcanet: Frequency channel attention
  networks,'' in \emph{Proceedings of the IEEE/CVF international conference on
  computer vision}, 2021, pp. 783--792.

\bibitem{csahinucc2022fractional}
F.~{\c{S}}ahinu{\c{c}} and A.~Ko{\c{c}}, ``Fractional fourier transform meets
  transformer encoder,'' \emph{IEEE Signal Processing Letters}, vol.~29, pp.
  2258--2262, 2022.

\bibitem{lee2021fnet}
J.~Lee-Thorp, J.~Ainslie, I.~Eckstein, and S.~Ontanon, ``Fnet: Mixing tokens
  with fourier transforms,'' \emph{arXiv preprint arXiv:2105.03824}, 2021.

\bibitem{wang2018packing}
Y.~Wang, C.~Xu, C.~Xu, and D.~Tao, ``Packing convolutional neural networks in
  the frequency domain,'' \emph{IEEE transactions on pattern analysis and
  machine intelligence}, vol.~41, no.~10, pp. 2495--2510, 2018.

\bibitem{pan2024dct}
H.~Pan, E.~Hamdan, X.~Zhu, K.~Biswas, A.~Cetin, and U.~Bagci, ``Dct-based
  decorrelated attention for vision transformers,'' \emph{arXiv preprint
  arXiv:2405.13901}, 2024.

\bibitem{he2023fourier}
Z.~He, M.~Yang, M.~Feng, J.~Yin, X.~Wang, J.~Leng, and Z.~Lin, ``Fourier
  transformer: Fast long range modeling by removing sequence redundancy with
  fft operator,'' \emph{arXiv preprint arXiv:2305.15099}, 2023.

\bibitem{li2023discrete}
X.~Li, Y.~Zhang, J.~Yuan, H.~Lu, and Y.~Zhu, ``Discrete cosin transformer:
  Image modeling from frequency domain,'' in \emph{Proceedings of the IEEE/CVF
  Winter Conference on Applications of Computer Vision}, 2023, pp. 5468--5478.

\bibitem{duan2022fourier}
H.~Duan, Y.~Liu, H.~Yan, Q.~He, Y.~He, and T.~Guan, ``Fourier vit: A
  multi-scale vision transformer with fourier transform for histopathological
  image classification,'' in \emph{2022 7th International Conference on
  Automation, Control and Robotics Engineering (CACRE)}.\hskip 1em plus 0.5em
  minus 0.4em\relax IEEE, 2022, pp. 189--193.

\bibitem{scribano2023dct}
C.~Scribano, G.~Franchini, M.~Prato, and M.~Bertogna, ``Dct-former: Efficient
  self-attention with discrete cosine transform,'' \emph{Journal of Scientific
  Computing}, vol.~94, no.~3, p.~67, 2023.

\bibitem{saha2000image}
S.~Saha, ``Image compression—from dct to wavelets: a review,'' \emph{XRDS:
  Crossroads, The ACM Magazine for Students}, vol.~6, no.~3, pp. 12--21, 2000.

\bibitem{pan2024multichannel}
H.~Pan, E.~Hamdan, X.~Zhu, S.~Atici, and A.~E. Cetin, ``Multichannel orthogonal
  transform-based perceptron layers for efficient resnets,'' \emph{IEEE
  Transactions on Neural Networks and Learning Systems}, 2024.

\bibitem{cai2023htnet}
J.~Cai, H.~Liu, R.~Ding, W.~Li, J.~Wu, and M.~Ban, ``Htnet: Human topology
  aware network for 3d human pose estimation,'' in \emph{ICASSP 2023-2023 IEEE
  International Conference on Acoustics, Speech and Signal Processing
  (ICASSP)}.\hskip 1em plus 0.5em minus 0.4em\relax IEEE, 2023, pp. 1--5.

\bibitem{cai2024disentangled}
Q.~Cai, X.~Hu, S.~Hou, L.~Yao, and Y.~Huang, ``Disentangled diffusion-based 3d
  human pose estimation with hierarchical spatial and temporal denoiser,'' in
  \emph{Proceedings of the AAAI conference on artificial intelligence},
  vol.~38, no.~2, 2024, pp. 882--890.

\bibitem{chartrand1969connectivity}
G.~Chartrand and M.~J. Stewart, ``The connectivity of line-graphs,''
  \emph{Mathematische Annalen}, vol. 182, pp. 170--174, 1969.

\bibitem{pavllo:videopose3d:2019}
D.~Pavllo, C.~Feichtenhofer, D.~Grangier, and M.~Auli, ``3d human pose
  estimation in video with temporal convolutions and semi-supervised
  training,'' in \emph{Conference on Computer Vision and Pattern Recognition
  (CVPR)}, 2019.

\bibitem{newell2016stacked}
A.~Newell, K.~Yang, and J.~Deng, ``Stacked hourglass networks for human pose
  estimation,'' in \emph{Computer Vision--ECCV 2016: 14th European Conference,
  Amsterdam, The Netherlands, October 11-14, 2016, Proceedings, Part VIII
  14}.\hskip 1em plus 0.5em minus 0.4em\relax Springer, 2016, pp. 483--499.

\bibitem{loshchilov2017decoupled}
I.~Loshchilov, ``Decoupled weight decay regularization,'' \emph{arXiv preprint
  arXiv:1711.05101}, 2017.

\bibitem{sun2019deep}
K.~Sun, B.~Xiao, D.~Liu, and J.~Wang, ``Deep high-resolution representation
  learning for human pose estimation,'' in \emph{Proceedings of the IEEE/CVF
  conference on computer vision and pattern recognition}, 2019, pp. 5693--5703.

\bibitem{shan2022p}
W.~Shan, Z.~Liu, X.~Zhang, S.~Wang, S.~Ma, and W.~Gao, ``P-stmo: Pre-trained
  spatial temporal many-to-one model for 3d human pose estimation,'' in
  \emph{European Conference on Computer Vision}.\hskip 1em plus 0.5em minus
  0.4em\relax Springer, 2022, pp. 461--478.

\bibitem{yu2023gla}
B.~X. Yu, Z.~Zhang, Y.~Liu, S.-h. Zhong, Y.~Liu, and C.~W. Chen, ``Gla-gcn:
  Global-local adaptive graph convolutional network for 3d human pose
  estimation from monocular video,'' in \emph{Proceedings of the IEEE/CVF
  International Conference on Computer Vision}, 2023, pp. 8818--8829.

\bibitem{zhong2023frame}
Y.~Zhong, G.~Yang, D.~Zhong, X.~Yang, and S.~Wang, ``Frame-padded multiscale
  transformer for monocular 3d human pose estimation,'' \emph{IEEE Transactions
  on Multimedia}, vol.~26, pp. 6191--6201, 2023.

\bibitem{xue2022boosting}
Y.~Xue, J.~Chen, X.~Gu, H.~Ma, and H.~Ma, ``Boosting monocular 3d human pose
  estimation with part aware attention,'' \emph{IEEE Transactions on Image
  Processing}, vol.~31, pp. 4278--4291, 2022.

\bibitem{zhang2024app}
J.~Zhang, M.~Liu, H.~Liu, G.~Wang, and W.~Li, ``App: Adaptive pose pooling for
  3d human pose estimation from videos,'' in \emph{Proceedings of the 32nd ACM
  International Conference on Multimedia}, 2024, pp. 1672--1681.

\bibitem{sun2024repose}
Z.~Sun, Y.~Liang, Z.~Ma, T.~Zhang, L.~Bao, G.~Li, and S.~He, ``Repose: 3d human
  pose estimation via spatio-temporal depth relational consistency,'' in
  \emph{European Conference on Computer Vision}.\hskip 1em plus 0.5em minus
  0.4em\relax Springer, 2024, pp. 309--325.

\end{thebibliography}
\end{document}